\PassOptionsToPackage{dvipsnames,table}{xcolor}
\PassOptionsToPackage{numbers,compress}{natbib}
\documentclass[11pt]{article}
\usepackage[final]{ropedia}
\tcbuselibrary{breakable,skins}

\usepackage[T1]{fontenc}
\usepackage[utf8]{inputenc}
\usepackage{amsmath,amssymb,amsfonts}
\usepackage{mathpazo}
\usepackage[xcharter,bigdelims,vvarbb]{newtxmath}
\usepackage[scaled]{helvet}
\usepackage[scaled=1.1]{zlmtt}
\usepackage{latexsym,pifont,placeins,booktabs,multirow,tabularx}
\usepackage{subcaption,arydshln,adjustbox,makecell,enumitem,rotating}
\usepackage{titletoc,comment,wrapfig,needspace,nicefrac,microtype}
\usepackage{siunitx,dblfloatfix,graphicx,cuted,capt-of,xspace}
\usepackage{afterpage,float}
\setlist[itemize]{leftmargin=2em}
\graphicspath{{figs/}}
\usepackage{tikz}
\usetikzlibrary{fadings}
\usepackage{array}
\usepackage{tcolorbox}
\usepackage{threeparttable}

\usepackage{amsmath,amsfonts,bm}

\def\eqref#1{equation~\ref{#1}}

\def\1{\bm{1}}

\DeclareMathAlphabet{\mathsfit}{\encodingdefault}{\sfdefault}{m}{sl}
\SetMathAlphabet{\mathsfit}{bold}{\encodingdefault}{\sfdefault}{bx}{n}

\newcommand{\projecttitle}{EgoXperience\xspace}
\newcommand{\projectsubtitle}{Learning World-Grounded Egocentric Representations from Human Experience\xspace}

\newcommand{\method}{\textsc{EgoXperience}\xspace}
\newcommand{\model}{\textsc{EgoX}\xspace}

\newcommand{\yes}{\ding{51}}
\newcommand{\no}{\ding{55}}

\newcommand{\pending}{\textcolor{red!65!black}{TBD}}

\definecolor{subcol}{HTML}{F7FFF2}
\colorlet{rankfirst}{ropediagreen}
\colorlet{ranksecond}{ropedialightgreen}
\colorlet{rankthird}{ropediaabstractgreen}
\definecolor{trainmark}{RGB}{0,180,0}
\definecolor{oomred}{RGB}{252,228,228}
\definecolor{catgray}{HTML}{E7F8DC}
\definecolor{altcolgreen}{HTML}{F0FFE8}
\definecolor{pretrainmark}{RGB}{197,90,17}
\definecolor{gaingreen}{RGB}{30,140,60}
\definecolor{lossred}{RGB}{200,30,30}
\definecolor{navyblue}{HTML}{4F8F35}

\colorlet{findingback}{ropedialightgreen}
\colorlet{findingframe}{ropediadeepgreen}
\colorlet{findingaccent}{ropediaaccent}

\firstpagelogos{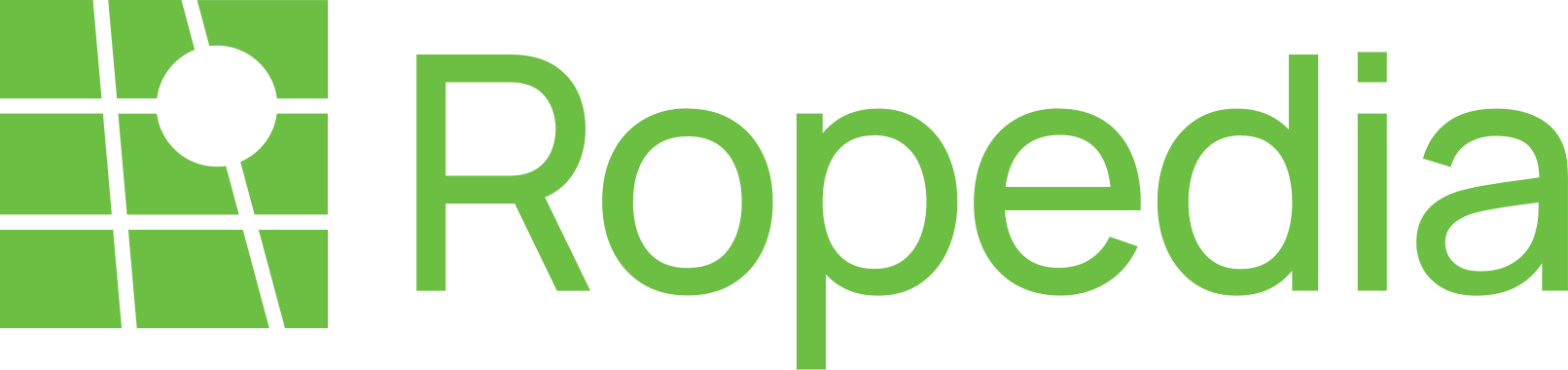}{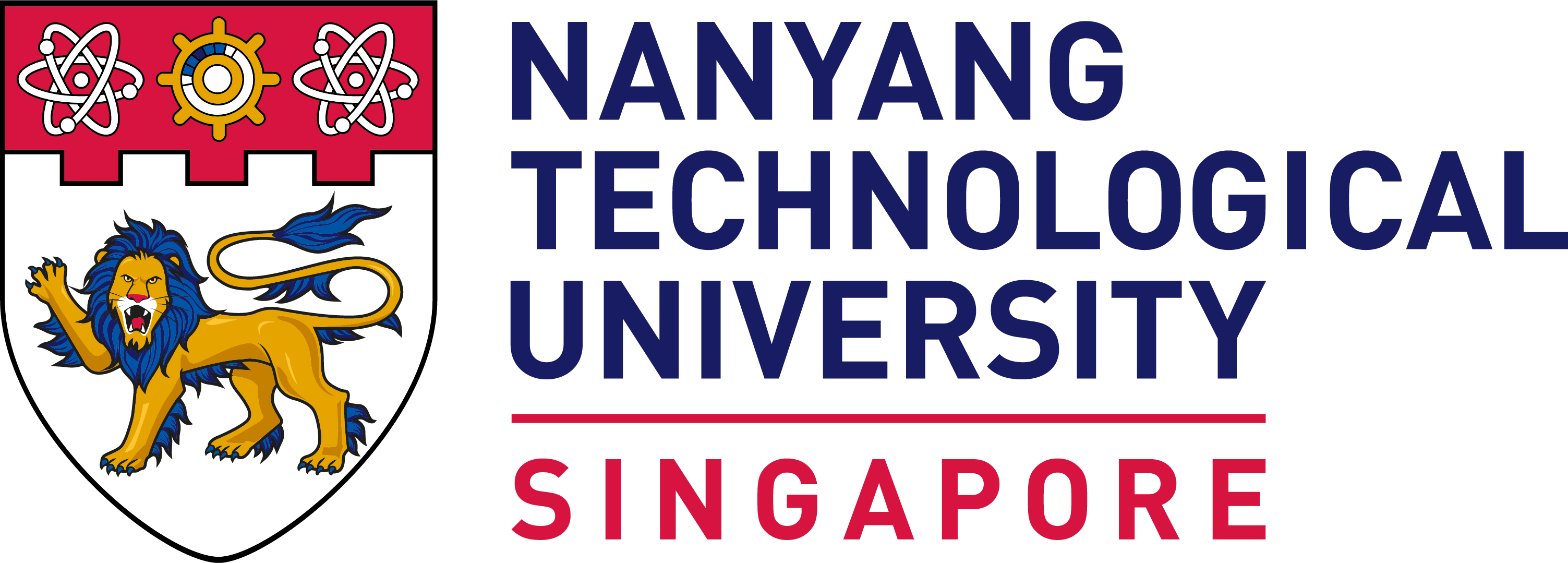}{}
\abstractlogo{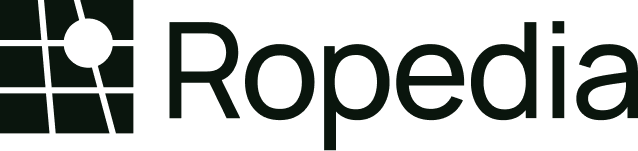}
\title{\projecttitle}
\subtitle{\projectsubtitle}

\newcommand{\authorfootnotes}{%
  \begingroup
  \renewcommand{\thefootnote}{}%
  \footnotetext{* Equal contribution. $\ddagger$ Project leader. \textsc{D} Data collection. $\dagger$ Corresponding author.}%
  \endgroup
  \addtocounter{footnote}{-1}%
}

\author{
  \parbox{\textwidth}{\centering
  \mbox{\RopediaAuthorName{Xinhao Ji}{1,3,*}}\quad
  \mbox{\RopediaAuthorName{Hui En Pang}{*}}\quad
  \mbox{\RopediaAuthorName{Hao Li}{1,*,$\ddagger$}}\quad
  \mbox{\RopediaAuthorName{Kairui Hu}{1}}\quad
  \mbox{\AuthorName{Jiner Wang}{1}}\quad
  \mbox{\RopediaAuthorName{Mingze Sun}{2}}\quad
  \mbox{\AuthorName{Ziang Cao}{1}}\quad
  \mbox{\RopediaAuthorName{Yalun Dai}{1}}\quad
  \mbox{\AuthorName{Thang Luu}{1,\textsc{D}}}\quad
  \mbox{\AuthorName{Qin Ma}{1,\textsc{D}}}\quad
  \mbox{\AuthorName{Hongyi Chen}{1,\textsc{D}}}\quad
  \mbox{\AuthorName{Yukang Cao}{1}}\quad
  \mbox{\textbf{Fangzhou Hong}\textsuperscript{\ropediamark}}\quad
  \mbox{\textbf{Zhaoxi Chen}\textsuperscript{\ropediamark}}\quad
  \mbox{\RopediaAuthorName{Ziwei Liu}{1,$\dagger$}}\\[1.0ex]
  {\normalfont\footnotesize
  \mbox{\ropediamark\ Ropedia}\quad
  \mbox{\textsuperscript{1}Nanyang Technological University}\quad
  \mbox{\textsuperscript{2}Tsinghua University}\quad
  \mbox{\textsuperscript{3}Peking University}}}
}

\begin{document}
\maketitle
\authorfootnotes
\begin{paperresources}
\makebox[\linewidth][l]{%
\paperresourceicon{\resourceprojecticon}
  {Project Page}
  {https://ropedia.github.io/EgoXperience/}
  {ropedia.github.io/EgoXperience}
\hfill
\paperresourceicon{\resourcegithubicon}
  {EgoXperience Code}
  {https://github.com/Ropedia/EgoXperience}
  {github.com/Ropedia/EgoXperience}
}%
\\[2pt]
\paperresourceicon{\resourcehficon}
  {Data \& Models}
  {https://huggingface.co/collections/ropedia-ai/egoxperience}
  {huggingface.co/collections/ropedia-ai/egoxperience}
\end{paperresources}
\vspace{6pt}
{\centering
\includegraphics[width=\textwidth]{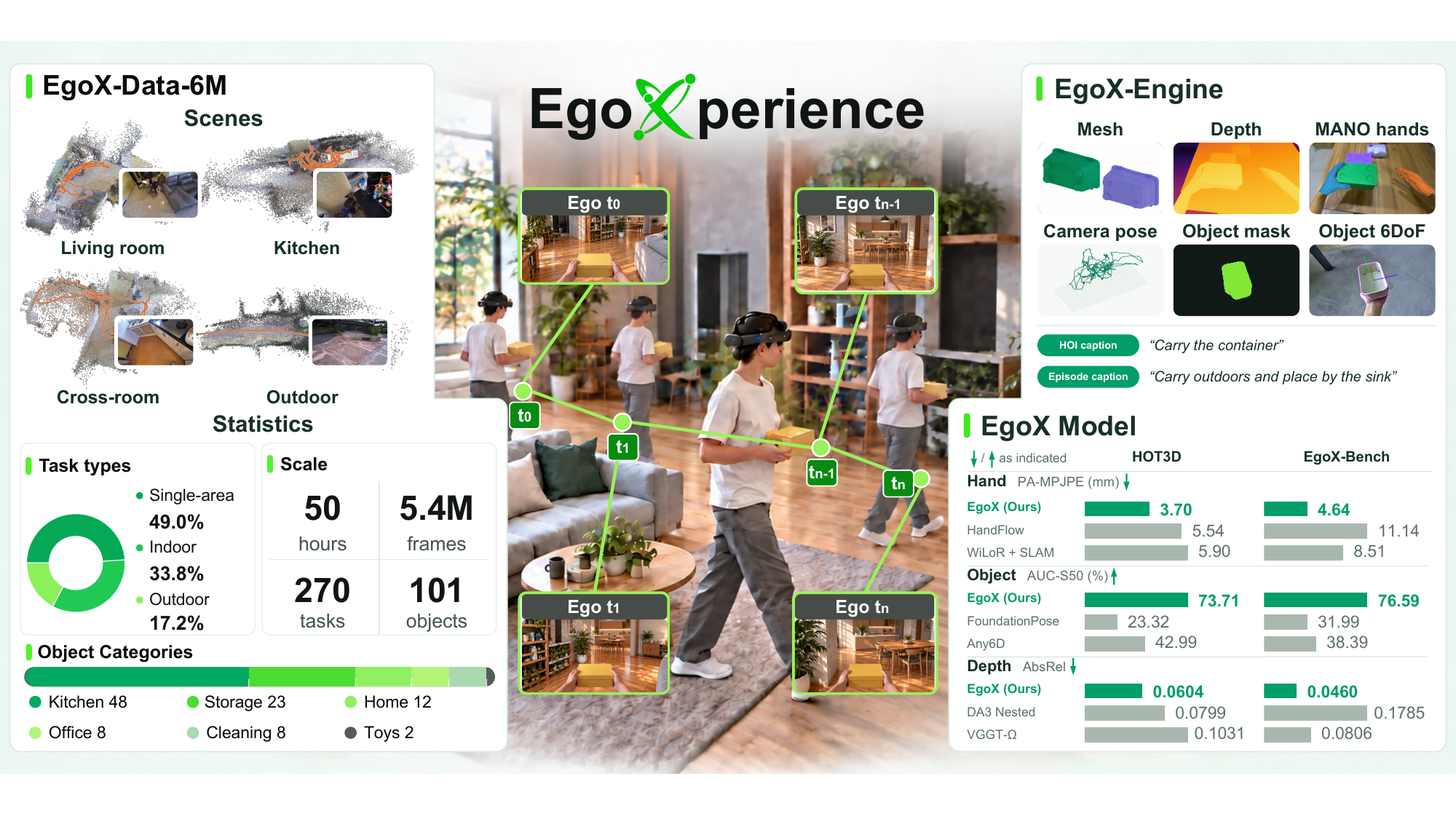}
\par\vspace{5pt}
\begin{minipage}{\textwidth}\footnotesize
\refstepcounter{figure}\label{fig:teaser}
Figure~\thefigure{} | \textbf{Overview of the EgoXperience Paradigm.}
Task-driven capture and EgoX-Engine produce EgoX-Data-6M, supporting joint reconstruction with EgoX.
\end{minipage}\par}
\vspace{6pt}

\begin{abstract}

Reconstructing human activity from egocentric video requires more than recovering individual 3D components:
the observer's motion, surrounding geometry, and hand-object interactions must fit together in a consistent world.
Achieving this coherence requires both joint reconstruction and aligned supervision, yet existing methods and datasets offer limited support for either.
We present \textbf{\method}, a full-stack paradigm that addresses these needs through data collection, automatic annotation, unified modeling, and benchmarking.
Central to EgoXperience is an automatic annotation engine, \textbf{\model-Engine}, that couples geometrically verified object tracking with SLAM-based world alignment, yielding consistent supervision for the camera, scene, hands, and objects.
This yields \textbf{\model-Data-6M}, a 50-hour dataset of task-driven manipulation and locomotion within and across indoor and outdoor environments, with second-level task captions.
We carefully select one hour across these activity categories as \textbf{\model-Bench} to evaluate joint reconstruction.
To translate this supervision into reconstruction, we introduce \textbf{\model}, trained with \model-Data-6M and complementary annotations from additional public datasets.
Given monocular egocentric video and first-frame object masks, our model achieves joint metric reconstruction of the camera, scene, hands, and objects in a shared world coordinate frame with a single forward pass. 
Extensive experiments demonstrate state-of-the-art performance across scene estimation and world-space hand-object pose recovery.
Against the strongest evaluated baselines, EgoX reduces camera, hand, and depth errors by 68.9\%, 65.6\%, and 42.9\%, and improves object pose scores by 28.0\%.

\end{abstract}
\section{Introduction}
\label{sec:intro}



Egocentric video offers a natural perspective on how humans and embodied agents perceive and interact with the world, requiring only a wearable camera without external camera setups.
By capturing the wearer's movement and close-range interactions, it provides visual cues for recovering camera pose, scene depth, hands, and objects.
However, existing methods~\citep{wang2025vggt, vggtomega2026,hand3r2026, reviv2026, mint2026} typically recover only a subset of these tightly coupled components.
Combining outputs from different methods often requires reconciling inconsistent scales and coordinate frames, complicating their use in downstream embodied AI applications.
This motivates a unified feed-forward framework that jointly recovers the camera, scene, hands, and objects within a single model.


Training such a unified model, however, requires large-scale egocentric data with joint annotations of the camera, scene, hands, and objects.
Although egocentric video is easy to record, obtaining reliable joint annotations is considerably more difficult due to frequent hand-object occlusions, rapid head motion, and motion blur.
Existing datasets~\citep{cao2026ace,hot3d2024,liu2022hoi4d,chao2021dexycb,hocap2024} have made progress toward this goal, often using calibrated multi-camera or motion-capture systems to support reliable annotations.
However, these systems require substantial setup and calibration, making collection difficult to scale across diverse real-world environments.
Scaling jointly annotated egocentric data therefore calls for a collection and annotation pipeline that reduces reliance on such specialized infrastructure.

To address these challenges, we present \method, a full-stack paradigm for egocentric 3D reconstruction, encompassing data collection, annotation, modeling, and benchmarking.
Central to its annotation pipeline is \textbf{\model-Engine}, which tracks target objects across video frames and uses pre-scanned object models and observed depth to estimate and verify their 3D poses.
SLAM-estimated camera poses place these object trajectories in a shared world coordinate system alongside scene geometry and hand annotations, providing geometrically aligned supervision.
Applying this engine to 50 hours of task-driven recordings spanning workspace manipulation, cross-room interactions, and outdoor loco-manipulation yields \textbf{\model-Data-6M}, further enriched with task captions at second-level temporal resolution.
To complement this dataset, we harmonize additional public interaction datasets~\citep{
hot3d2024,liu2022hoi4d,kwon2021h2o,hocap2024,
keypointtransformer2022,chao2021dexycb} under a common training formulation, aligning their geometric conventions while retaining their partial annotations.
We also establish \textbf{\model-Bench} from one hour of held-out recordings, enabling joint evaluation of camera, scene depth, hand, and object reconstruction across all capture settings.

To translate this joint supervision into joint reconstruction, we introduce \textbf{\model}.
Given monocular egocentric video and first-frame object masks, \model jointly reconstructs the camera, scene depth, hands, and objects at metric scale in a shared world coordinate system, all in a single forward pass.
During training, we supervise each component and additionally use the predicted camera poses to transform object and hand predictions into normalized world coordinates, where we supervise their positions and orientations.
This ties camera, hand, and object estimation together, encouraging geometrically consistent reconstruction.
Evaluated on four benchmarks, \model achieves state-of-the-art performance across scene estimation and world-space hand–object pose recovery. It ranks first in depth, hand, and object metrics across all four benchmarks and achieves the best camera accuracy on HOI4D. Relative to the strongest competing methods, EgoX reduces camera nATE by 68.9\% on HOI4D, hand W-MPJPE
and depth AbsRel by 65.6\% and 42.9\% on EgoX-Bench,
respectively, and improves object AUC-S$_{50}$ by 28.0\% on HOT3D.

\section{Related Work}
\label{sec:related}

\subsection{Scene and entity reconstruction}
\label{sec:related:geometry}

DUSt3R~\citep{wang2024dust3r} predicts pairwise pointmaps, while
VGGT~\citep{wang2025vggt} jointly estimates cameras and scene
geometry. CUT3R~\citep{wang2025cut3r} introduces recurrent scene
memory, and VGGT-$\Omega$~\citep{vggtomega2026} reduces inter-frame
attention through scene registers. Dynamic-scene methods recover
moving geometry through world-space point tracking or temporally
conditioned point prediction~\citep{zhang2025monst3r,st4rtrack2025,
dpm2025,vdpm2026,d4rt2026,cao2026crowdmogen}. Entity-aware models incorporate humans,
instances, or hands into reconstruction: Human3R~\citep{human3r2025}
jointly estimates human meshes, cameras, and scenes; HSImul3R~\citep{cao2026hsimul3r} enables simulator-ready joint reconstruction of humans and scenes;
IGGT~\citep{iggt2026} couples geometry with instance segmentation;
and HGGT~\citep{hggt2026} estimates MANO parameters and cameras from
uncalibrated views. 
EgoX extends this line of work to moving hands and manipulated objects, jointly reconstructing their motion, camera poses, and depth in a shared metric frame.

\subsection{Hand-object reconstruction}
\label{sec:related:hands}
\label{sec:related:hoi}
\label{sec:related:entity}
\label{sec:related:pose}
\label{sec:related:scale}

HaMeR~\citep{pavlakos2024hamer} and WiLoR~\citep{potamias2025wilor}
estimate MANO hand meshes from images. For world-space hand motion,
HaWoR~\citep{hawor2025} combines hand reconstruction with SLAM and
motion infilling, while Dyn-HaMR~\citep{dynhamr2025} uses camera
tracking, motion priors, and test-time optimization.
Hand3R~\citep{hand3r2026}, MINT~\citep{mint2026}, and
ReViV~\citep{reviv2026} couple hand reconstruction with cameras or
scene geometry; ReViV additionally predicts depth, body motion, and
gaze. EgoX further incorporates manipulated-object motion into the
same feed-forward reconstruction.

Keypoint Transformer~\citep{keypointtransformer2022} and
HOPformer~\citep{hopformer2026} estimate hand--object interactions
without recovering a world-camera trajectory. HOLD~\citep{fan2024hold}
optimizes hand and object representations from video, while
ForeHOI~\citep{forehoi2026} completes object masks and shapes before
geometric pose estimation. EgoGrasp~\citep{egograsp2026} reconstructs
hands and objects in separate stages, and WHOLE~\citep{whole2026}
requires object templates, a metric camera trajectory, and test-time
guidance. Related object-pose systems use CAD models, reference
images, or RGB-D observations~\citep{labbe2022megapose,
wen2024foundationpose,any6d2025}. In contrast, EgoX jointly estimates camera poses, depth, hand geometry, and rigid-object poses from RGB observations and optional first-frame object masks.

\subsection{Egocentric datasets and supervision}
\label{sec:related:data}

HO-3D~\citep{hampali2020ho3d}, DexYCB~\citep{chao2021dexycb},
H2O~\citep{kwon2021h2o}, H2O-3D~\citep{keypointtransformer2022},
HOI4D~\citep{liu2022hoi4d}, ARCTIC~\citep{fan2023arctic},
HO-Cap~\citep{hocap2024}, and HOT3D~\citep{hot3d2024} provide
hand-object annotations under different capture and geometric
conventions. Recent datasets broaden scene coverage:
SHOW3D~\citep{show3d2026} records indoor and outdoor interactions
with a mobile ego--exo rig, Aria Digital Twin~\citep{adt2023}
provides scanned scenes and world-space trajectories, and
HD-EPIC~\citep{hdepic2025} captures unscripted activities with
digital twins. Large corpora such as Ego4D~\citep{grauman2022ego4d},
EPIC-KITCHENS~\citep{damen2022epickitchens}, and
EgoDex~\citep{egodex2025} provide broader human experience but more
limited aligned geometric supervision. 
ACE-Data-0~\citep{cao2026ace} builds an ambient capture engine for calibrated and synchronized multimodal data collection in a controlled environment.
EgoX-Data-6M complements
these resources with aligned camera, depth, hand, and object
annotations across workspace manipulation, cross-area interaction,
and outdoor locomotion.


\section{The \method Paradigm}
\label{sec:paradigm}

Our \method unifies data capture, annotation and model learning for joint metric reconstruction of the camera, scene depth, hands, and objects from egocentric videos.
As illustrated in Fig.~\ref{fig:teaser}, this paradigm comprises \model-Engine for data processing and annotation (Sec.~\ref{sec:engine}), \model-Data-6M with 50 hours of task-driven recordings spanning manipulation and locomotion (Sec.~\ref{sec:data}), and \model for joint reconstruction from monocular video (Sec.~\ref{sec:model}).

\definecolor{CapabilityYes}{HTML}{00B050}
\definecolor{CapabilityNo}{HTML}{E53935}
\definecolor{EGGTCapabilityGray}{HTML}{E2F3E8}

\providecommand{\yes}{\ding{51}}
\providecommand{\no}{\ding{55}}

\begin{table*}[t]
\centering
\begingroup

\renewcommand{\yes}{\textcolor{CapabilityYes}{\ding{51}}}
\renewcommand{\no}{\textcolor{CapabilityNo}{\ding{55}}}

\caption{\textbf{Comparison of Egocentric Datasets.}
We compare data scale, capture settings, and annotation coverage.
Obj. counts objects with 3D models.}
\label{tab:datasets}

\scriptsize
\setlength{\tabcolsep}{2.2pt}
\renewcommand{\arraystretch}{1.08}

\begin{adjustbox}{width=\linewidth}
\begin{tabular}{@{}lrr*{9}{c}@{}}
\toprule
& & &
\multicolumn{3}{c}{Capture setting} &
\multicolumn{6}{c}{Annotations} \\
\cmidrule(lr){4-6}
\cmidrule(lr){7-12}

Dataset & Hours & Obj.
& Single-area & Cross-area & Outdoor
& 3D Hand & 6DoF & Mesh & Depth & Mask & Caption \\
\midrule

H2O~\citep{kwon2021h2o}
& 1.1 & 8
& \yes & \no & \no
& \yes & \yes & \yes & \yes & \no & \no \\

HOI4D~\citep{liu2022hoi4d}
& 22 & 800
& \yes & \no & \no
& \yes & \yes & \yes & \yes & \yes & \no \\

ARCTIC~\citep{fan2023arctic}
& 2.2 & 11
& \yes & \no & \no
& \yes & \yes & \yes & \no & \no & \no \\

Ego-Exo4D~\citep{grauman2024egoexo4d}
& 221 & 0
& \yes & \no & \yes
& \yes & \no & \no & \no & \no & \yes \\

EgoDex~\citep{egodex2025}
& 829 & 0
& \yes & \no & \no
& \yes & \no & \no & \no & \no & \yes \\

HOT3D~\citep{hot3d2024}
& 13.9 & 33
& \yes & \no & \no
& \yes & \yes & \yes & \no & \yes & \no \\

HO-Cap~\citep{hocap2024}
& 0.7 & 64
& \yes & \no & \no
& \yes & \yes & \yes & \yes & \yes & \no \\

EgoVerse~\citep{egoverse2026}
& 1362 & 0
& \yes & \no & \no
& \yes & \no & \no & \no & \no & \yes \\

SHOW3D~\citep{show3d2026}
& 20 & 21
& \yes & \yes & \yes
& \yes & \yes & \yes & \no & \no & \yes \\

ACE-Data-0~\citep{cao2026ace}
& 150 & 50
& \yes & \yes & \no
& \yes & \yes & \yes & \no & \yes & \yes \\

\midrule
\rowcolor{EGGTCapabilityGray}
\textbf{EgoX-Data-6M}
& \textbf{50} & \textbf{101}
& \yes & \yes & \yes
& \yes & \yes & \yes & \yes & \yes & \yes \\

\bottomrule
\end{tabular}
\end{adjustbox}

\endgroup
\end{table*}

\subsection{\model-Engine}
\label{sec:engine}

\begin{figure*}[t]
\centering
\includegraphics[width=\textwidth,keepaspectratio]{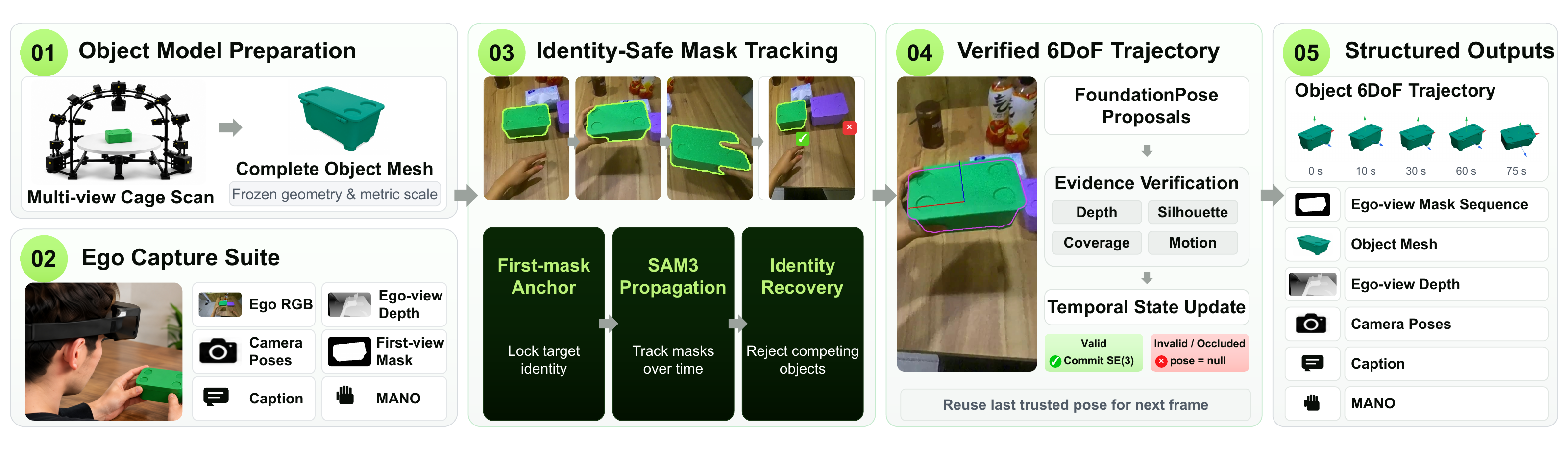}
\caption{\textbf{Overview of EgoX-Engine.}
The pipeline converts egocentric recordings into structured
multimodal annotations, including aligned geometry and
temporally localized task captions.}
\label{fig:egoxperience_pipeline}
\end{figure*}

Learning to jointly reconstruct egocentric activity requires annotations that describe the camera, scene, hands, and objects in relation to one another.
Obtaining these annotations is challenging because the observer and the interacting entities move simultaneously:
a change in image position may arise from camera motion, entity motion, or both.
\model-Engine addresses this ambiguity by using stereo depth to recover geometry at metric scale and SLAM-estimated camera poses to align different annotations throughout each recording.
This allows hand-object motion to be expressed relative to the surrounding scene as the observer moves.
Fig.~\ref{fig:egoxperience_pipeline} summarizes the overall pipeline.

\paragraph{Capture and geometric reference.}

We record with the HOMIE Gen2 head-mounted rig~\citep{homie2026}, using its associated processing pipeline to estimate metric depth from the synchronized, calibrated front stereo pair and recover
camera trajectories via SLAM from its four calibrated fisheye cameras.
For each recording, we use the SLAM coordinate system as a fixed session frame and transform camera-relative scene, hand, and object estimates into this frame.

\begin{figure*}[t]
\centering
\includegraphics[
    width=\textwidth,
    keepaspectratio
]{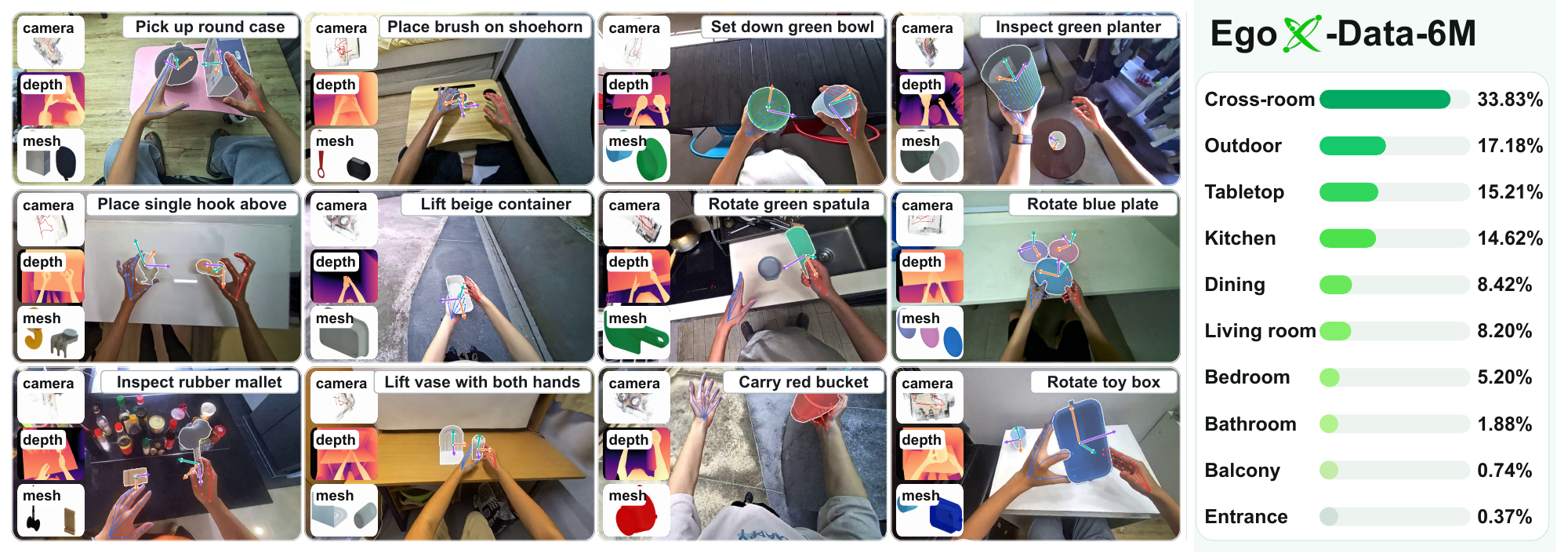}
\caption{\textbf{Overview of EgoX-Data-6M.}
Examples and statistics summarize capture settings,
task coverage, and aligned multimodal annotations.}
\label{fig:dataset}
\end{figure*}
\paragraph{Object geometry and motion.}

Object annotation requires both a geometric reference for each instance and its changing pose during interaction.
We construct metric object meshes from 31 synchronized cameras mounted on a calibrated dome. 
SAM~3~\citep{sam3_2025} then extracts object masks from the captured views, which are used for mesh reconstruction with MV-SAM3D~\citep{mvsam3d2026}.
To recover the object's physical dimensions, we put an AprilTag~\citep{olson2011apriltag} board inside the dome to provide the scale reference.

Given these metric object references, the engine recovers each object's motion in the egocentric recording.
Specifically, 
we first identify the target object using a first-frame mask and then specify the interaction interval over which to track it.
Using the masked RGB image and its corresponding stereo depth, we estimate the initial pose of the pre-scanned object model in the camera frame.
From this initial pose, we apply FoundationPose~\citep{wen2024foundationpose} to track the object throughout the specified interval, producing a camera-relative pose sequence.
We verify this sequence by checking whether the object model placed at the estimated pose agrees with the observed depth.
Sequences that pass verification are transformed through the SLAM-estimated camera poses into the session frame, aligning object motion with the scene and hand annotations.

\paragraph{Scene and hand annotations.}

The scene and hand annotations used in this alignment are derived from the same calibrated stereo observations.
Specifically, we reconstruct scene geometry by back-projecting stereo depth and recover 3D hand keypoints through cross-view triangulation.
We use these keypoints to constrain MANO fitting, with additional reprojection, depth, and temporal terms to refine the estimated hand geometry and motion.
Hand shape parameters are shared across frames within each recording, while hand pose, position, and orientation are estimated independently for each frame.
Finally, we transform the fitted hands and scene points through the camera trajectory into the session frame, placing them in the same metric coordinate system as the tracked objects.

\subsection{\model-Data-6M}
\label{sec:data}

Given \model-Engine, we apply it to annotate task-driven egocentric captures, constructing \model-Data-6M, which comprises 50 hours of video, 101 rigid objects, and 270 task scripts.
The recordings cover activities ranging from manipulation within a single workspace to interactions across rooms and loco-manipulation in outdoor environments.
The engine provides geometrically aligned camera, scene, hand, and object annotations across these settings, supporting joint reconstruction under both local manipulation and extended movement.
Fig.~\ref{fig:dataset} illustrates the activity coverage and accompanying annotations.

\paragraph{Task-driven collection.}

We organize data collection around task scripts, each specifying a manipulation goal, target objects, and
a capture setting.
Specifically, workspace scripts focus on local object manipulation, whereas cross-room and outdoor scripts incorporate movement between interaction locations.

\paragraph{Annotation coverage.}

We provide geometric annotations in a session-fixed metric frame, allowing camera motion, scene structure, and hand-object interactions to be related throughout each recording.
Additionally, we include second-level task captions to describe the ongoing activity.
Tab.~\ref{tab:datasets} compares our dataset with existing egocentric resources.
Compared with existing datasets, our \model-Data-6M brings together workspace manipulation, cross-room interactions, and outdoor locomotion with camera, scene, hand, and object annotations in a shared metric frame, supporting joint reconstruction across these activity settings.

\paragraph{Benchmark and complementary training data.}

To evaluate all reconstruction components on the same recordings, we reserve a quality-controlled, one-hour subset spanning the three activity settings as \model-Bench.
This subset is used exclusively for joint evaluation of camera, scene, hand, and object reconstruction.

To broaden the training data beyond our own recordings, we incorporate additional public datasets.
Using these datasets together requires accounting for differences in both geometric conventions and available annotations. 
We therefore convert their geometric labels to a common coordinate and scale convention and supervise the model using the labels available in each dataset.
This allows recordings with partial annotations to contribute to training without requiring every dataset to provide camera, scene, hand, and object labels.
We also clean erroneous annotations and align depth with hand and object geometry to reduce conflicting supervision (Appendix~\ref{app:depth}).
Sec.~\ref{sec:model} explains how the model learns from this combined supervision; the data mixture is detailed in Appendix~\ref{app:training}.

\subsection{\model}
\label{sec:model}

\begin{figure*}[t]
\centering
\includegraphics[width=\textwidth]{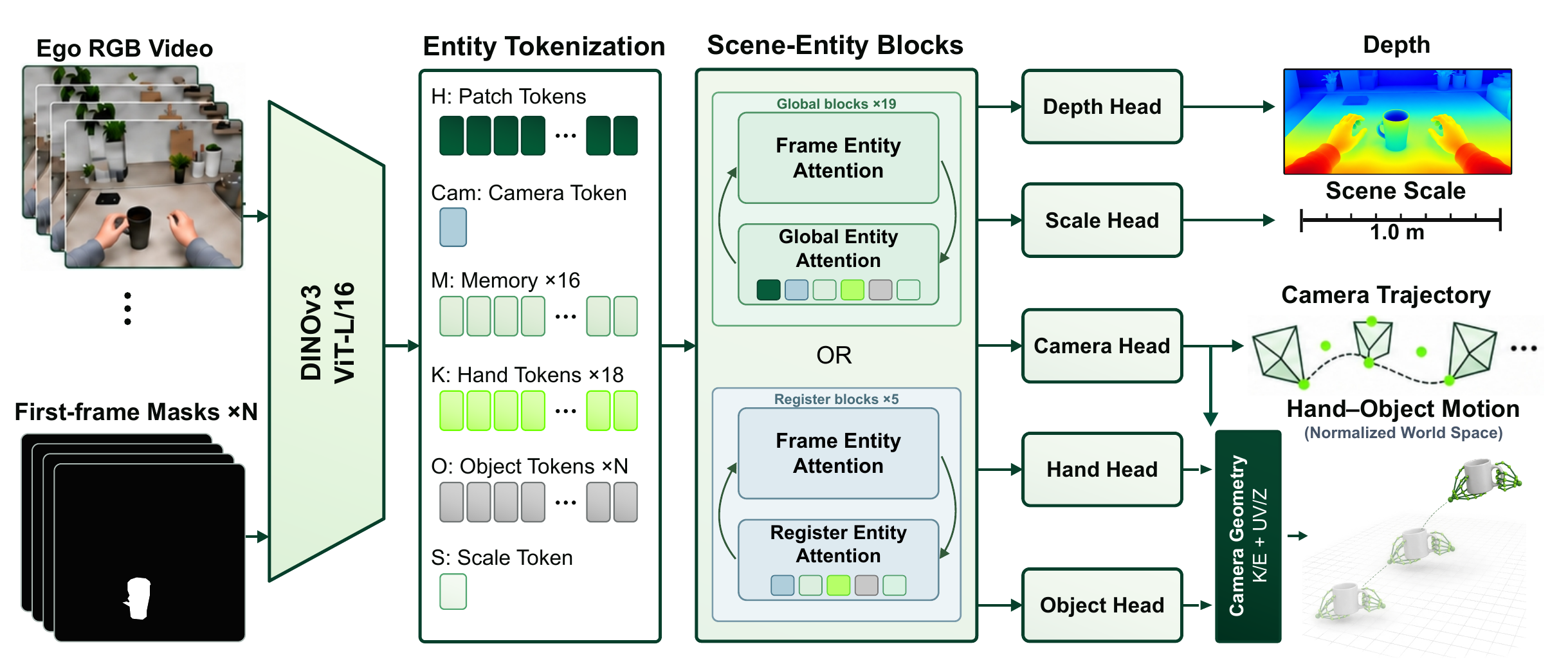}
\caption{\textbf{Overview of EgoX.}
Shared scene and entity features support joint reconstruction,
while predicted cameras and scale place all outputs in a
common metric world frame.}
\label{fig:architecture}
\end{figure*}

To translate the aligned supervision into joint reconstruction, we propose \model, which is capable of predicting the camera, scene, and motions of both hands and objects within a single model.
The central challenge to achieve this goal is that the predictions must agree geometrically as both the viewpoint and interaction change.
As illustrated in Fig.~\ref{fig:architecture}, we address this through a joint scene and entity representation, followed by explicit composition through the predicted camera poses.
Supervision on the resulting hand and object motions then ties their reconstruction to camera estimation.

\paragraph{Problem formulation.}

Given monocular observations $\mathcal I=\{I_t\}_{t=1}^{T}$ and optional first-frame object masks $\mathcal B=\{B_{1,j}\}_{j=1}^{N}$ identifying the target objects, \model jointly predicts camera intrinsics and poses, dense depth, two-hand MANO states, object origins and orientations, and a positive metric scale in a single forward pass.
We express all reconstructed outputs in \(\mathcal W\), the coordinate system of the first-frame camera.
For those complementary datasets without ground-truth object masks, we adopt the pipeline to reconstruct only cameras, scene geometry, and hands.

\begin{figure*}[t]
\centering
\includegraphics[width=\textwidth]{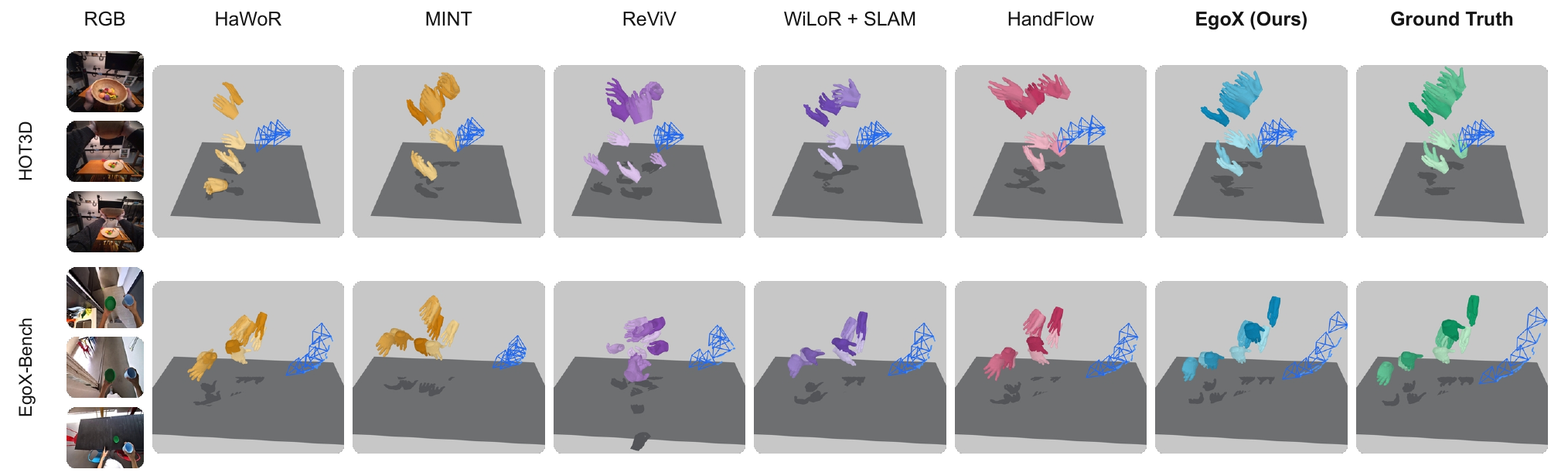}
\caption{\textbf{World-Space Hand and Camera Reconstruction.}
Qualitative comparisons on HOT3D and EgoX-Bench.
Predictions and ground truth are shown in a common world frame.}
\label{fig:hand_camera_comparison}
\end{figure*}

\begin{figure*}[t]
\centering
\includegraphics[width=\textwidth]{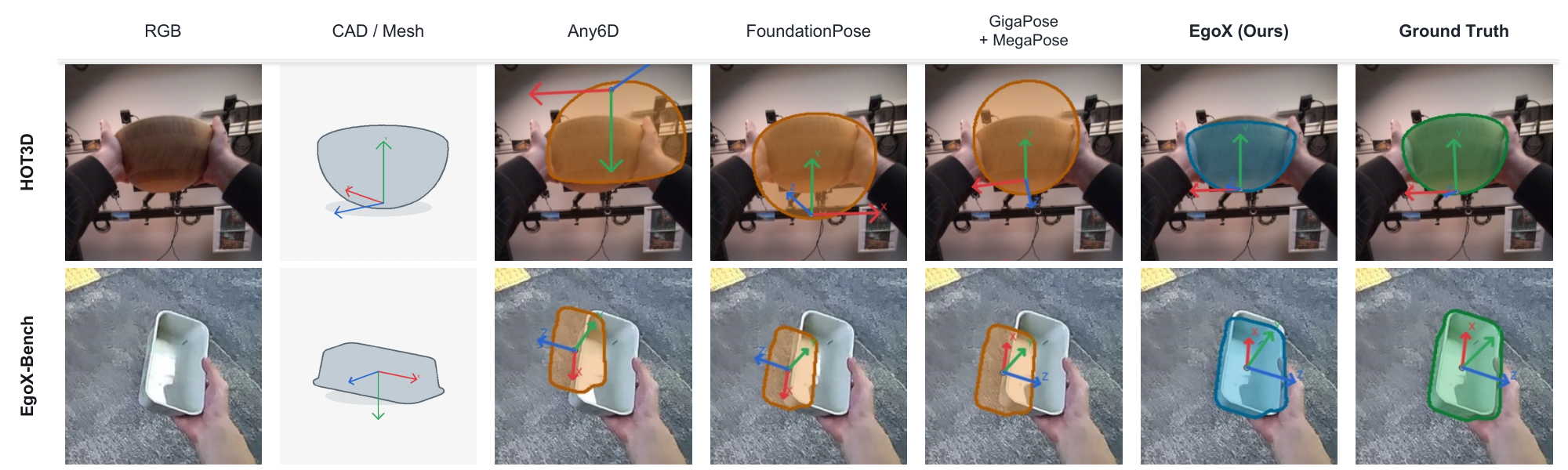}
\caption{\textbf{Camera-Frame Object Pose Reconstruction.}
Qualitative comparisons of object 6-DoF predictions
with ground truth on HOT3D and EgoX-Bench.}
\label{fig:object_pose_comparison}
\end{figure*}

\begin{figure*}[t]
\centering
\includegraphics[width=\textwidth]{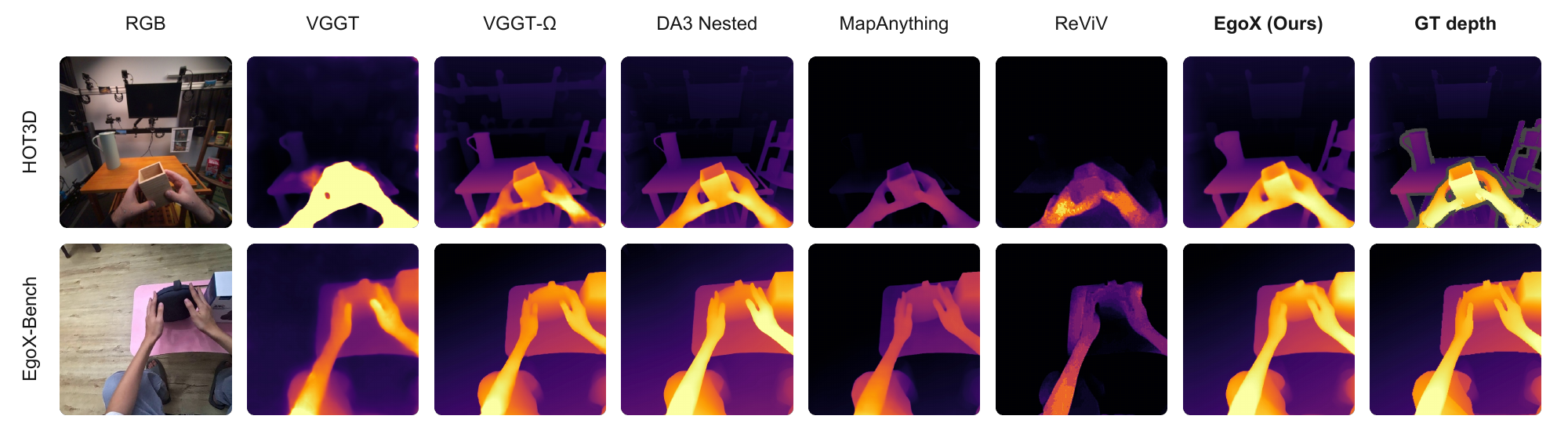}
\caption{\textbf{Depth Reconstruction.}
GT: refined official depth. VGGT/VGGT-$\Omega$: one scale fitted
to DA3 Nested depth per sequence; others: predicted metric scale.}
\label{fig:depth_comparison}
\end{figure*}

\afterpage{%
  \afterpage{%
    \definecolor{CapabilityYes}{HTML}{00B050}
\definecolor{CapabilityNo}{HTML}{E53935}
\definecolor{EGGTCapabilityGray}{HTML}{E2F3E8}

\begin{table}[H]
\centering
\caption{\textbf{Reconstruction Outputs and Inference Structure.}
Checkmarks denote supported outputs; inference is classified
as end-to-end or multi-stage.}
\label{tab:capabilities}
\begingroup
\scriptsize
\renewcommand{\yes}{\textcolor{CapabilityYes}{\ding{51}}}
\renewcommand{\no}{\textcolor{CapabilityNo}{\ding{55}}}
\setlength{\tabcolsep}{2.2pt}
\renewcommand{\arraystretch}{1.05}

\begin{adjustbox}{width=\linewidth}
\begin{tabular}{@{}l*{5}{c}l@{}}
\toprule
\multirow[c]{2}{*}{Method}
& \multicolumn{5}{c}{Outputs}
& \multirow[c]{2}{*}{Inference} \\
\cmidrule(lr){2-6}
& Camera & Depth & Hands & Object 6DoF & World-space & \\
\midrule
VGGT-$\Omega$~\citep{vggtomega2026} & \yes & \yes & \no & \no & \no & End-to-end \\
MapAnything~\citep{mapanything2025} & \yes & \yes & \no & \no & \yes & End-to-end \\
DA3 Nested~\citep{da3_2025} & \yes & \yes & \no & \no & \yes & End-to-end \\
\addlinespace[1.5pt]
HaWoR~\citep{hawor2025} & \yes & \no & \yes & \no & \yes & Multi-stage \\
Hand3R~\citep{hand3r2026} & \yes & \yes & \yes & \no & \yes & End-to-end \\
MINT~\citep{mint2026} & \yes & \no & \yes & \no & \yes & End-to-end \\
ReViV~\citep{reviv2026} & \yes & \yes & \yes & \no & \yes & End-to-end \\
HandFlow~\citep{handflow2026} & \yes & \no & \yes & \no & \yes & End-to-end \\
\addlinespace[1.5pt]
FoundationPose~\citep{wen2024foundationpose} & \no & \no & \no & \yes & \no & Multi-stage \\
Any6D~\citep{any6d2025} & \no & \no & \no & \yes & \no & Multi-stage \\
GigaPose + MegaPose~\citep{gigapose2024,labbe2022megapose} & \no & \no & \no & \yes & \no & Multi-stage \\
HOPformer~\citep{hopformer2026} & \no & \no & \yes & \yes & \no & End-to-end \\
WHOLE~\citep{whole2026} & \no & \no & \yes & \yes & \yes & Multi-stage \\
\midrule
\rowcolor{EGGTCapabilityGray}
\textbf{EgoX} & \yes & \yes & \yes & \yes & \yes & End-to-end \\
\bottomrule
\end{tabular}
\end{adjustbox}

\endgroup
\end{table}
  }%
}

\paragraph{Joint scene and entity representation.}

To jointly estimate cameras, scene geometry, and hand-object motion, we encode them with a shared Transformer.
Starting from image patch features extracted by DINOv3~\citep{dinov3_2025}, we introduce camera, hand, and object tokens in each view, using distinct left/right embeddings to identify the hands and first-frame mask descriptors to associate target objects across views.
These tokens gather image evidence through within-view attention and exchange information across observations through cross-view attention, forming a joint representation for reconstruction.
Architectural details are provided in Appendix~\ref{app:training}.

\paragraph{Reconstruction in a shared metric frame.}

From the joint representation, we reconstruct scene geometry and hand-object motion in a common reference frame $\mathcal W$, defined by the first-frame camera coordinate system.
To combine predictions across views, we estimate camera intrinsics $\hat K_t$ and extrinsics $(\hat R_t,\hat\tau_t)$ mapping $\mathcal W$ to camera $t$.
These parameters allow us to recover scene points, wrist positions, and object origins from their image coordinates and predicted depths.
Specifically, for a homogeneous image position $\tilde u=(u_x,u_y,1)^\top$ and normalized depth $z$, its position in $\mathcal W$ is:
\begin{equation}
    \mathcal U_t(\tilde u,z)
    =
    \hat R_t^\top
    \left(z\hat K_t^{-1}\tilde u-\hat\tau_t\right).
    \label{eq:shared_composition}
\end{equation}
This shared mapping ties scene and entity positions to the same camera estimates.
For hand and object orientations, we similarly use inverse camera rotation $\hat R_t^\top$ to transform camera-frame hand and object orientations into $\mathcal W$, keeping MANO joint rotations local and expressing object rotations relative to their first-frame orientations.

The reconstruction in $\mathcal W$ obtained above still uses normalized depths and translations.
To recover metric scale, we apply a shared ScaleHead to predict one positive factor $\hat s$ for the sequence, scaling scene points and camera, wrist, and object positions consistently:
$\hat p^{\mathcal W}=\hat s\,\mathcal U_t(\tilde u,z)$.
For hands, this adjusts only the wrist positions while preserving the metric dimensions of the MANO meshes.

\paragraph{Geometrically coupled supervision.}

Building on this shared reconstruction, we supervise not only individual predictions but also the resulting hand-object geometry in $\mathcal W$.
To compute this geometric supervision, we transform hand and object predictions through the estimated cameras and compare the reconstructed geometry with ground truth.
The resulting loss backpropagates through the camera transformations, updating both entity and camera predictions.
We combine these supervision terms into the training objective:
\begin{equation}
    \mathcal L
    =
    \mathcal L_{\mathrm{local}}
    + \mathcal L_{\mathrm{dense}}
    + \mathcal L_{\mathrm{coupled}}
    + \mathcal L_{\mathrm{scale}},
    \label{eq:complete_objective}
\end{equation}
where $\mathcal L_{\mathrm{local}}$ supervises camera and view-local entity predictions, $\mathcal L_{\mathrm{dense}}$ supervises depth and scene points, $\mathcal L_{\mathrm{coupled}}$ supervises the composed hand-object geometry, and $\mathcal L_{\mathrm{scale}}$ supervises metric scale.
To accommodate varying annotation coverage across datasets, we apply each loss only where its required labels are available, allowing fully and partially annotated data to contribute to the same objective.
Loss definitions and training settings are provided in Appendix~\ref{app:training}.

\section{Experiments}
\label{sec:experiments}
\label{sec:results}



We evaluate camera, depth, hand, and object reconstruction on HOT3D,
EgoX-Bench, H2O, and HOI4D using identical 150-observation clips
across methods. Evaluation sequences are excluded from EgoX training.
EgoX uses RGB and first-frame object masks; object baselines may
additionally require estimated depth, intrinsics, per-frame masks,
or metric CAD assets. World-space hand metrics use estimated cameras;
object poses are evaluated in camera coordinates. No method receives
GT camera poses or scale in the main comparisons.
Appendix~\ref{app:evaluation} details the evaluation protocols.

\subsection{Evaluation scope}
\label{sec:evaluation_scope}

The evaluated methods cover different subsets of the reconstruction
problem. Table~\ref{tab:capabilities} summarizes their outputs and
inference structure. Subsequent comparisons therefore evaluate each
method only on the outputs it provides under the matched protocol
described above. EgoX is the only end-to-end method in this comparison
that jointly reconstructs cameras, depth, hands, and object 6-DoF
poses in a shared world frame.

\definecolor{EGGTResultBlue}{HTML}{E2F3E8}

\begin{table}[!t]
\centering
\caption{\textbf{Hand Reconstruction on Four Benchmarks.}
PA-, WA-, and W-MPJPE are reported in millimetres.
World-space errors use each method's estimated camera trajectory.}
\label{tab:hand_results}

\scriptsize
\setlength{\tabcolsep}{3.2pt}
\renewcommand{\arraystretch}{1.00}
\providecommand{\pending}{\texttt{TBD}}
\providecommand{\notreported}{\textemdash}

\begin{adjustbox}{max width=\linewidth}
\begin{tabular}{@{}l*{12}{c}@{}}
\toprule
& \multicolumn{3}{c}{HOT3D}
& \multicolumn{3}{c}{EgoX-Bench}
& \multicolumn{3}{c}{HOI4D}
& \multicolumn{3}{c}{H2O} \\
\cmidrule(lr){2-4}
\cmidrule(lr){5-7}
\cmidrule(lr){8-10}
\cmidrule(lr){11-13}
Method
& \shortstack{PA-\\MPJPE} $\downarrow$ & \shortstack{WA-\\MPJPE} $\downarrow$ & \shortstack{W-\\MPJPE} $\downarrow$
& \shortstack{PA-\\MPJPE} $\downarrow$ & \shortstack{WA-\\MPJPE} $\downarrow$ & \shortstack{W-\\MPJPE} $\downarrow$
& \shortstack{PA-\\MPJPE} $\downarrow$ & \shortstack{WA-\\MPJPE} $\downarrow$ & \shortstack{W-\\MPJPE} $\downarrow$
& \shortstack{PA-\\MPJPE} $\downarrow$ & \shortstack{WA-\\MPJPE} $\downarrow$ & \shortstack{W-\\MPJPE} $\downarrow$ \\
\midrule

HaWoR
& 6.06 & 29.80 & \underline{56.30}
& 10.98 & 125.73 & 363.47
& 18.61 & 50.01 & 110.96
& 5.68 & 25.83 & 64.62 \\

MINT
& 9.94 & 37.76 & 85.85
& 10.19 & 149.91 & 467.65
& \underline{17.94} & \underline{43.64} & \underline{108.76}
& 5.74 & 28.38 & 50.04 \\

ReViV
& 14.01 & 49.90 & 109.02
& 23.89 & 111.48 & 501.29
& 25.18 & 83.40 & 354.39
& 8.69 & 23.40 & 45.22 \\

WiLoR + SLAM
& 5.90 & \underline{25.52} & 58.55
& \underline{8.51} & \underline{63.03} & \underline{328.81}
& 19.41 & 47.66 & 132.94
& \underline{5.11} & 21.78 & 74.95 \\

HandFlow
& \underline{5.54} & 59.01 & 119.48
& 11.14 & 75.76 & 375.11
& 19.91 & 57.02 & 120.26
& 7.07 & \underline{19.00} & \underline{34.63} \\

\midrule
\textbf{EgoX (Ours)}
& \cellcolor{EGGTResultBlue}\textbf{3.70}
& \cellcolor{EGGTResultBlue}\textbf{15.54}
& \cellcolor{EGGTResultBlue}\textbf{38.89}
& \cellcolor{EGGTResultBlue}\textbf{4.64}
& \cellcolor{EGGTResultBlue}\textbf{53.81}
& \cellcolor{EGGTResultBlue}\textbf{113.24}
& \cellcolor{EGGTResultBlue}\textbf{3.32}
& \cellcolor{EGGTResultBlue}\textbf{15.12}
& \cellcolor{EGGTResultBlue}\textbf{42.67}
& \cellcolor{EGGTResultBlue}\textbf{4.97}
& \cellcolor{EGGTResultBlue}\textbf{11.41}
& \cellcolor{EGGTResultBlue}\textbf{23.95}
\\

\bottomrule
\end{tabular}
\end{adjustbox}
\par
\addvspace{\floatsep}

\caption{\textbf{Camera-Frame Object Pose Reconstruction.}
ADD/ADD-S: mm; AUC: \%. Extra inputs: $\hat D$/$\hat K$,
DA3 Nested depth/intrinsics; $M$, per-frame masks; $A$, GT metric CAD.}
\label{tab:object_results}

\scriptsize
\setlength{\tabcolsep}{3.2pt}
\renewcommand{\arraystretch}{1.00}
\providecommand{\pending}{\texttt{TBD}}
\providecommand{\notreported}{\textemdash}

\begin{adjustbox}{max width=\linewidth}
\begin{tabular}{@{}l*{12}{c}@{}}
\toprule
& \multicolumn{3}{c}{HOT3D}
& \multicolumn{3}{c}{EgoX-Bench}
& \multicolumn{3}{c}{HOI4D}
& \multicolumn{3}{c}{H2O} \\
\cmidrule(lr){2-4}
\cmidrule(lr){5-7}
\cmidrule(lr){8-10}
\cmidrule(lr){11-13}
Method
& ADD $\downarrow$ & ADD-S $\downarrow$ & AUC-S$_{50}$ $\uparrow$
& ADD $\downarrow$ & ADD-S $\downarrow$ & AUC-S$_{50}$ $\uparrow$
& ADD $\downarrow$ & ADD-S $\downarrow$ & AUC-S$_{50}$ $\uparrow$
& ADD $\downarrow$ & ADD-S $\downarrow$ & AUC-S$_{50}$ $\uparrow$ \\
\midrule

Any6D$^{\hat D,\hat K,M}$
& \underline{67.06} & 44.05 & 42.99
& 88.25 & 41.41 & 38.39
& \underline{67.01} & \underline{53.26} & \underline{34.61}
& \underline{72.75} & 36.57 & 38.37 \\

FoundationPose$^{\hat D,\hat K,A}$
& 92.49 & 59.87 & 23.32
& 85.26 & 55.04 & 31.99
& 94.76 & 73.02 & 13.39
& 91.91 & \underline{30.97} & \underline{46.03} \\

GigaPose + MegaPose$^{\hat K,A,M}$
& 71.92 & \underline{32.42} & \underline{57.59}
& \underline{69.96} & \underline{26.87} & \underline{59.80}
& 97.94 & 88.68 & 3.91
& 89.08 & 53.08 & 29.14 \\

\midrule
\textbf{EgoX (Ours)}
& \cellcolor{EGGTResultBlue}\textbf{28.00}
& \cellcolor{EGGTResultBlue}\textbf{14.03}
& \cellcolor{EGGTResultBlue}\textbf{73.71}
& \cellcolor{EGGTResultBlue}\textbf{27.11}
& \cellcolor{EGGTResultBlue}\textbf{12.41}
& \cellcolor{EGGTResultBlue}\textbf{76.59}
& \cellcolor{EGGTResultBlue}\textbf{44.55}
& \cellcolor{EGGTResultBlue}\textbf{21.11}
& \cellcolor{EGGTResultBlue}\textbf{59.08}
& \cellcolor{EGGTResultBlue}\textbf{22.10}
& \cellcolor{EGGTResultBlue}\textbf{10.29}
& \cellcolor{EGGTResultBlue}\textbf{80.42} \\

\bottomrule
\end{tabular}
\end{adjustbox}
\end{table}

\definecolor{EGGTResultBlue}{HTML}{E2F3E8}

\begin{table}[!t]
\centering
\caption{\textbf{Camera and Depth Reconstruction.}
RRE: degrees; $\delta_{1.25}$: \%.
VGGT/VGGT-$\Omega$: one scale fitted to DA3 Nested depth
per sequence; others: predicted metric scale.}
\label{tab:scene_results}

\scriptsize
\setlength{\tabcolsep}{3.4pt}
\renewcommand{\arraystretch}{1.00}
\providecommand{\pending}{\texttt{TBD}}
\providecommand{\notreported}{\textemdash}

\begin{adjustbox}{max width=\linewidth}
\begin{tabular}{@{}l*{16}{c}@{}}
\toprule
& \multicolumn{4}{c}{HOT3D}
& \multicolumn{4}{c}{EgoX-Bench}
& \multicolumn{4}{c}{HOI4D}
& \multicolumn{4}{c}{H2O} \\
\cmidrule(lr){2-5}
\cmidrule(lr){6-9}
\cmidrule(lr){10-13}
\cmidrule(lr){14-17}

& \multicolumn{2}{c}{Camera}
& \multicolumn{2}{c}{Depth$^\dagger$}
& \multicolumn{2}{c}{Camera}
& \multicolumn{2}{c}{Depth$^\dagger$}
& \multicolumn{2}{c}{Camera}
& \multicolumn{2}{c}{Depth$^\dagger$}
& \multicolumn{2}{c}{Camera}
& \multicolumn{2}{c}{Depth$^\dagger$} \\
\cmidrule(lr){2-3}
\cmidrule(lr){4-5}
\cmidrule(lr){6-7}
\cmidrule(lr){8-9}
\cmidrule(lr){10-11}
\cmidrule(lr){12-13}
\cmidrule(lr){14-15}
\cmidrule(lr){16-17}

Method
& nATE $\downarrow$ & RRE ($^\circ$) $\downarrow$
& AbsRel $\downarrow$ & $\delta_{1.25}$ $\uparrow$
& nATE $\downarrow$ & RRE ($^\circ$) $\downarrow$
& AbsRel $\downarrow$ & $\delta_{1.25}$ $\uparrow$
& nATE $\downarrow$ & RRE ($^\circ$) $\downarrow$
& AbsRel $\downarrow$ & $\delta_{1.25}$ $\uparrow$
& nATE $\downarrow$ & RRE ($^\circ$) $\downarrow$
& AbsRel $\downarrow$ & $\delta_{1.25}$ $\uparrow$ \\
\midrule

VGGT
& 0.0064 & 0.64 & 0.2182 & 69.40
& 0.0684 & 2.86 & 0.2877 & 66.53
& 0.0310 & \underline{1.50} & 0.0676 & 98.60
& 0.0036 & \underline{0.27} & \underline{0.0726} & 94.36 \\

VGGT-$\Omega$
& \cellcolor{EGGTResultBlue}\textbf{0.0018}
& \cellcolor{EGGTResultBlue}\textbf{0.16}
& 0.1031 & 92.20
& \cellcolor{EGGTResultBlue}\textbf{0.0071}
& \cellcolor{EGGTResultBlue}\textbf{0.61}
& \underline{0.0806} & \underline{91.57}
& \underline{0.0212} & 1.56 & \underline{0.0558} & \underline{99.68}
& \underline{0.0030} & \underline{0.27} & 0.0784 & \underline{95.46} \\

CUT3R
& 0.0449 & 3.75 & 0.2748 & 69.19
& 0.0756 & 6.29 & 0.4317 & 54.07
& 0.0455 & 5.08 & 0.0647 & 97.05
& 0.0120 & 1.56 & 0.0887 & 93.09 \\

Any4D
& 0.0267 & 2.24 & 0.2899 & 70.23
& 0.0604 & 4.55 & 0.4161 & 48.15
& 0.0273 & 1.91 & 0.0757 & 95.80
& 0.0083 & 0.67 & 0.1075 & 87.26 \\

DA3 Nested
& \underline{0.0036} & \underline{0.23}
& \underline{0.0799} & \underline{94.40}
& \underline{0.0135} & \underline{0.92} & 0.1785 & 82.29
& 0.0307 & 1.54 & 0.0746 & 97.34
& \cellcolor{EGGTResultBlue}\textbf{0.0026}
& \cellcolor{EGGTResultBlue}\textbf{0.26}
& 0.0823 & 92.64 \\

MapAnything
& 0.0321 & 19.40 & 0.3182 & 64.20
& 0.0283 & 2.11 & 0.1429 & 87.12
& 0.0220 & 1.63 & 0.0727 & 95.36
& 0.0071 & 0.51 & 0.1009 & 92.01 \\

ReViV
& 0.0285 & 1.98 & 0.2558 & 61.19
& 0.0350 & 3.81 & 0.3972 & 53.60
& 0.0231 & 2.06 & 0.2075 & 60.42
& 0.0072 & 0.84 & 0.2562 & 53.07 \\

\midrule

\textbf{EgoX (Ours)}
& 0.0059 & 0.88
& \cellcolor{EGGTResultBlue}\textbf{0.0604}
& \cellcolor{EGGTResultBlue}\textbf{96.21}
& 0.0295 & 2.21 
& \cellcolor{EGGTResultBlue}\textbf{0.0460}
& \cellcolor{EGGTResultBlue}\textbf{97.37}
& \cellcolor{EGGTResultBlue}\textbf{0.0066}
& \cellcolor{EGGTResultBlue}\textbf{0.78}
& \cellcolor{EGGTResultBlue}\textbf{0.0312}
& \cellcolor{EGGTResultBlue}\textbf{99.98}
& 0.0035 & 0.33 
& \cellcolor{EGGTResultBlue}\textbf{0.0700}
& \cellcolor{EGGTResultBlue}\textbf{99.70} \\

\bottomrule
\end{tabular}
\end{adjustbox}

\end{table}

\subsection{Main results}
\label{sec:results:hot3d}
\label{sec:results:object}

\paragraph{World-space hands.}
EgoX achieves the lowest PA-, WA-, and W-MPJPE on all four benchmarks
(Table~\ref{tab:hand_results}). On HOT3D, it reduces W-MPJPE from
56.30 to 38.89\,mm relative to HaWoR, the strongest baseline on this
metric, despite HaWoR's adaptive DROID-SLAM and Metric3D-based
scale~\citep{hawor2025}. On EgoX-Bench, W-MPJPE falls from
328.81\,mm for WiLoR + SLAM to 113.24\,mm.
Figure~\ref{fig:hand_camera_comparison} illustrates improved
world-space placement: baseline hands recover plausible local hand poses but exhibit large deviations from the ground-truth positions in world coordinates, whereas EgoX closely matches the ground truth.


\paragraph{Object 6-DoF.}
EgoX achieves the best ADD, ADD-S, and AUC-S$_{50}$ on all four
benchmarks (Table~\ref{tab:object_results}). On EgoX-Bench, its
AUC-S$_{50}$ is 76.59, compared with 59.80 for GigaPose + MegaPose,
the strongest competing result, although that pipeline additionally
uses metric CAD assets and per-frame masks. EgoX also outperforms
Any6D on HOI4D, where Any6D receives estimated depth, intrinsics,
and per-frame masks. 
These comparisons show more accurate camera-frame poses with
fewer auxiliary inputs (Appendix~\ref{app:evaluation}).
Figure~\ref{fig:object_pose_comparison} shows larger translation or
rotation offsets for the object baselines, while EgoX more closely
matches the ground-truth pose.

\paragraph{Camera motion.}
Compared with ReViV, which also reconstructs cameras, depth, and
hands, EgoX achieves lower nATE and RRE on all four benchmarks
(Table~\ref{tab:scene_results}). On HOI4D, EgoX also leads the
geometry baselines: nATE is 0.0066 versus 0.0212 for
VGGT-$\Omega$, and RRE is $0.78^\circ$ versus $1.50^\circ$
for VGGT. VGGT-$\Omega$ leads on HOT3D and EgoX-Bench, while DA3 Nested
leads on H2O. VGGT-$\Omega$'s camera advantage may reflect its
large-scale geometric training on diverse static and dynamic
scenes~\citep{vggtomega2026}.

\paragraph{Depth reconstruction.}
EgoX achieves the lowest AbsRel and highest $\delta_{1.25}$ on all
four benchmarks (Table~\ref{tab:scene_results}). On EgoX-Bench, AbsRel is 0.0460 versus 0.3972 for ReViV and 0.0806 for VGGT-$\Omega$;
$\delta_{1.25}$ is 97.37 versus 91.57 for VGGT-$\Omega$.
EgoX obtains these results with its own
predicted metric scale, while VGGT-$\Omega$ uses scale estimated
from DA3 Nested depth (Section~\ref{sec:setup}). 
The results
support accurate metric depth alongside hand and object
reconstruction. Figure~\ref{fig:depth_comparison} shows that EgoX better matches the refined ground-truth depth.



\subsection{Ablation studies}
\label{sec:ablations}

\paragraph{Metric scale prediction.}
We remove ScaleHead and use oracle scale (G) to convert the ablated model’s predictions to metric units. Despite this advantage, W-MPJPE increases from 38.89
to 40.67\,mm and AUC-S$_{50}$ decreases from 73.71 to 60.87
(Table~\ref{tab:scale_ablation}), showing that ScaleHead benefits joint
reconstruction beyond final rescaling.

\paragraph{Dense-depth prediction.}
We remove DepthHead while retaining the remaining architecture and
predicted scale. W-MPJPE increases to 67.53\,mm and AUC-S$_{50}$
decreases to 41.09 (Table~\ref{tab:depth_ablation}), showing that explicit depth prediction improves
hand and object reconstruction through the shared representation.
\definecolor{EGGTResultBlue}{HTML}{E2F3E8}
\begin{table}[t]
\centering

\begin{minipage}[t]{0.48\textwidth}
\centering
\caption{\textbf{ScaleHead Ablation.}
Results on HOT3D with predicted (P) or oracle (G) scale.}
\label{tab:scale_ablation}

\scriptsize
\setlength{\tabcolsep}{3pt}
\renewcommand{\arraystretch}{1.00}
\begin{tabular}{@{}lcccc@{}}
\toprule
Setting & $s$
& W $\downarrow$
& AUC-S$_{50}$ $\uparrow$
& AbsRel $\downarrow$ \\
\midrule
w/ head & P
& \cellcolor{EGGTResultBlue}\textbf{38.89}
& \cellcolor{EGGTResultBlue}\textbf{73.71}
& \cellcolor{EGGTResultBlue}\textbf{0.0604} \\
w/o head & G
& \underline{40.67}
& \underline{60.87}
& \underline{0.0620} \\
\bottomrule
\end{tabular}
\end{minipage}
\hfill
\begin{minipage}[t]{0.48\textwidth}
\centering
\caption{\textbf{DepthHead Ablation.}
Results on HOT3D with predicted scale in both settings.}
\label{tab:depth_ablation}

\scriptsize
\setlength{\tabcolsep}{3pt}
\renewcommand{\arraystretch}{1.00}
\begin{tabular}{@{}lcccc@{}}
\toprule
Setting & $s$
& W $\downarrow$
& AUC-S$_{50}$ $\uparrow$
& AbsRel $\downarrow$ \\
\midrule
w/ head & P
& \cellcolor{EGGTResultBlue}\textbf{38.89}
& \cellcolor{EGGTResultBlue}\textbf{73.71}
& 0.0604 \\
w/o head & P
& \underline{67.53}
& \underline{41.09}
& \textit{N/A} \\
\bottomrule
\end{tabular}
\end{minipage}

\end{table}

\section{Conclusion}
\label{sec:limitations}


We present EgoXperience, a unified approach to learning
world-grounded egocentric representations through task-driven
capture, geometrically aligned annotation, and joint reconstruction.
EgoX-Engine provides aligned multimodal annotations for
the 50-hour EgoX-Data-6M, from which we curate the held-out
one-hour EgoX-Bench.
Trained on these annotations and harmonized public datasets,
EgoX jointly reconstructs cameras, dense depth, hands,
and rigid-object 6-DoF poses in a common metric world frame.
Experiments on EgoX-Bench and public benchmarks demonstrate
strong performance in hand and object reconstruction
and depth estimation.
Our current formulation assumes rigid objects and does not
predict their meshes.
Future work will explore object mesh prediction and
multiview inputs.

\clearpage
\section*{Acknowledgements}
We thank Hongyi Chen, Yuancheng Wang, Changyi Ye, Qin Ma, Thang Luu,
Wendi Fu, Qiankuan Tao, Sayang Mu, Rui Duan, Zetong Wang, and Yaobo Liu
for their contributions to data collection.

{
\small
\bibliography{references}

@InProceedings{hot3d2024,
    author    = {Banerjee, Prithviraj and Shkodrani, Sindi and Moulon, Pierre and Hampali, Shreyas and Han, Shangchen and Zhang, Fan and Zhang, Linguang and Fountain, Jade and Miller, Edward and Basol, Selen and Newcombe, Richard and Wang, Robert and Engel, Jakob Julian and Hodan, Tomas},
    title     = {HOT3D: Hand and Object Tracking in 3D from Egocentric Multi-View Videos},
    booktitle = {Proceedings of the IEEE/CVF Conference on Computer Vision and Pattern Recognition (CVPR)},
    month     = {June},
    year      = {2025},
    pages     = {7061-7071}
}

@inproceedings{hopformer2026,
  title     = {Towards in-the-wild Egocentric 3D Hand-Object Pose Estimation},
  author    = {Bansal, Siddhant and Zhu, Zhifan and Tripathi, Shashank and
               Zhao, Jiahe and Black, Michael J. and Damen, Dima},
  booktitle = {European Conference on Computer Vision (ECCV)},
  year      = {2026}
}

@article{cao2026ace,
  title   = {ACE-Data-0: Human-Centric Ambient Capture as Embodied Data Engine},
  author  = {Cao, Yukang and Xie, Haozhe and Wen, Beichen and Yao, Runmao and
             Liu, Yinghao and Huang, Yue and Liao, Zhichao and Wang, Yunxiang and
             Liu, Haiheng and Tian, Xingshun and Su, Dawei and Zhuo, Long and
             Tao, Dacheng and Wang, Xiaogang and Pan, Liang and Liu, Ziwei},
  journal = {arXiv preprint arXiv:2607.28625},
  year    = {2026}
}

@article{physx,
  title={PhysX-3D: Physical-Grounded 3D Asset Generation},
  author={Cao, Ziang and Chen, Zhaoxi and Pan, Liang and Liu, Ziwei},
  journal={arXiv preprint arXiv:2507.12465},
  year={2025}
}

@article{physxanything,
  title={PhysX-Anything: Simulation-Ready Physical 3D Assets from Single Image},
  author={Cao, Ziang and Hong, Fangzhou and Chen, Zhaoxi and Pan, Liang and Liu, Ziwei},
  journal={arXiv preprint arXiv:2511.13648},
  year={2025}
}

@misc{physxomni,
      title={PhysX-Omni: Unified Simulation-Ready Physical 3D Generation for Rigid, Deformable, and Articulated Objects}, 
      author={Ziang Cao and Yinghao Liu and Haitian Li and Runmao Yao and Fangzhou Hong and Zhaoxi Chen and Liang Pan and Ziwei Liu},
      year={2026},
      eprint={2605.21572},
      archivePrefix={arXiv},
      primaryClass={cs.CV},
      url={https://arxiv.org/abs/2605.21572}, 
}

@misc{sam3_2025,
      title={SAM 3: Segment Anything with Concepts},
      author={Nicolas Carion and Laura Gustafson and Yuan-Ting Hu and Shoubhik Debnath and Ronghang Hu and Didac Suris and Chaitanya Ryali and Kalyan Vasudev Alwala and Haitham Khedr and Andrew Huang and Jie Lei and Tengyu Ma and Baishan Guo and Arpit Kalla and Markus Marks and Joseph Greer and Meng Wang and Peize Sun and Roman Rädle and Triantafyllos Afouras and Effrosyni Mavroudi and Katherine Xu and Tsung-Han Wu and Yu Zhou and Liliane Momeni and Rishi Hazra and Shuangrui Ding and Sagar Vaze and Francois Porcher and Feng Li and Siyuan Li and Aishwarya Kamath and Ho Kei Cheng and Piotr Dollár and Nikhila Ravi and Kate Saenko and Pengchuan Zhang and Christoph Feichtenhofer},
      year={2025},
      eprint={2511.16719},
      archivePrefix={arXiv},
      primaryClass={cs.CV},
      url={https://arxiv.org/abs/2511.16719},
}

@INPROCEEDINGS{chao2021dexycb,
  author    = {Yu-Wei Chao and Wei Yang and Yu Xiang and Pavlo Molchanov and Ankur Handa and Jonathan Tremblay and Yashraj S. Narang and Karl {Van Wyk} and Umar Iqbal and Stan Birchfield and Jan Kautz and Dieter Fox},
  booktitle = {IEEE/CVF Conference on Computer Vision and Pattern Recognition (CVPR)},
  title     = {{DexYCB}: A Benchmark for Capturing Hand Grasping of Objects},
  year      = {2021},
}

@misc{forehoi2026,
      title={ForeHOI: Feed-forward 3D Object Reconstruction from Daily Hand-Object Interaction Videos}, 
      author={Yuantao Chen and Jiahao Chang and Chongjie Ye and Chaoran Zhang and Zhaojie Fang and Chenghong Li and Xiaoguang Han},
      year={2026},
      eprint={2602.06226},
      archivePrefix={arXiv},
      primaryClass={cs.CV},
      url={https://arxiv.org/abs/2602.06226}, 
}

@article{human3r2025,
    title={Human3R: Everyone Everywhere All at Once},
    author={Chen, Yue and Chen, Xingyu and Xue, Yuxuan and Chen, Anpei and Xiu, Yuliang and Gerard, Pons-Moll},
    journal={arXiv preprint arXiv:2510.06219},
    year={2025}
    }

@article{damen2022epickitchens,
  title   = {Rescaling Egocentric Vision: Collection, Pipeline and Challenges for {EPIC-KITCHENS-100}},
  author  = {Damen, Dima and Doughty, Hazel and Farinella, Giovanni Maria and Furnari, Antonino and Kazakos, Evangelos and Ma, Jian and Moltisanti, Davide and Munro, Jonathan and Perrett, Toby and Price, Will and Wray, Michael},
  journal = {International Journal of Computer Vision},
  year    = {2022}
}

@inproceedings{fan2023arctic,
  title = {{ARCTIC}: A Dataset for Dexterous Bimanual Hand-Object Manipulation},
  author = {Fan, Zicong and Taheri, Omid and Tzionas, Dimitrios and Kocabas, Muhammed and Kaufmann, Manuel and Black, Michael J. and Hilliges, Otmar},
  booktitle = {Proceedings IEEE Conference on Computer Vision and Pattern Recognition (CVPR)},
  year = {2023}
}

@inproceedings{fan2024hold,
  title={{HOLD}: Category-agnostic 3d reconstruction of interacting hands and objects from video},
  author={Fan, Zicong and Parelli, Maria and Kadoglou, Maria Eleni and Kocabas, Muhammed and Chen, Xu and Black, Michael J and Hilliges, Otmar},
  booktitle={Proceedings of the IEEE/CVF Conference on Computer Vision and Pattern Recognition},
  pages={494--504},
  year={2024}
}

@inproceedings{st4rtrack2025,
  title={St4RTrack: Simultaneous 4D Reconstruction and Tracking in the World},
  author={Feng, Haiwen and Zhang, Junyi and Wang, Qianqian and Ye, Yufei and Yu, Pengcheng and Black, Michael J. and Darrell, Trevor and Kanazawa, Angjoo},
  booktitle={Proceedings of the IEEE/CVF International Conference on Computer Vision},
  year={2025}
}

@misc{egograsp2026,
      title={EgoGrasp: World-Space Hand-Object Interaction Estimation from Egocentric Videos}, 
      author={Hongming Fu and Wenjia Wang and Xiaozhen Qiao and Rolandos Alexandros Potamias and Taku Komura and Shuo Yang and Zheng Liu and Bo Zhao},
      year={2026},
      eprint={2601.01050},
      archivePrefix={arXiv},
      primaryClass={cs.CV},
      url={https://arxiv.org/abs/2601.01050}, 
}

@inproceedings{grauman2022ego4d,
      title={Ego4D: Around the World in 3,000 Hours of Egocentric Video}, 
      author={Kristen Grauman and Andrew Westbury and Eugene Byrne and Zachary Chavis and Antonino Furnari and Rohit Girdhar and Jackson Hamburger and Hao Jiang and Miao Liu and Xingyu Liu and Miguel Martin and Tushar Nagarajan and Ilija Radosavovic and Santhosh Kumar Ramakrishnan and Fiona Ryan and Jayant Sharma and Michael Wray and Mengmeng Xu and Eric Zhongcong Xu and Chen Zhao and Siddhant Bansal and Dhruv Batra and Vincent Cartillier and Sean Crane and Tien Do and Morrie Doulaty and Akshay Erapalli and Christoph Feichtenhofer and Adriano Fragomeni and Qichen Fu and Abrham Gebreselasie and Cristina Gonzalez and James Hillis and Xuhua Huang and Yifei Huang and Wenqi Jia and Weslie Khoo and Jachym Kolar and Satwik Kottur and Anurag Kumar and Federico Landini and Chao Li and Yanghao Li and Zhenqiang Li and Karttikeya Mangalam and Raghava Modhugu and Jonathan Munro and Tullie Murrell and Takumi Nishiyasu and Will Price and Paola Ruiz Puentes and Merey Ramazanova and Leda Sari and Kiran Somasundaram and Audrey Southerland and Yusuke Sugano and Ruijie Tao and Minh Vo and Yuchen Wang and Xindi Wu and Takuma Yagi and Ziwei Zhao and Yunyi Zhu and Pablo Arbelaez and David Crandall and Dima Damen and Giovanni Maria Farinella and Christian Fuegen and Bernard Ghanem and Vamsi Krishna Ithapu and C. V. Jawahar and Hanbyul Joo and Kris Kitani and Haizhou Li and Richard Newcombe and Aude Oliva and Hyun Soo Park and James M. Rehg and Yoichi Sato and Jianbo Shi and Mike Zheng Shou and Antonio Torralba and Lorenzo Torresani and Mingfei Yan and Jitendra Malik},
      year={2022},
    booktitle={Proceedings of the IEEE/CVF Conference on Computer Vision and Pattern Recognition},
    pages={18995-19012},
}

@inproceedings{grauman2024egoexo4d,
      title={Ego-Exo4D: Understanding Skilled Human Activity from First- and Third-Person Perspectives}, 
      author={Kristen Grauman and Andrew Westbury and Lorenzo Torresani and Kris Kitani and Jitendra Malik and Triantafyllos Afouras and Kumar Ashutosh and Vijay Baiyya and Siddhant Bansal and Bikram Boote and Eugene Byrne and Zach Chavis and Joya Chen and Feng Cheng and Fu-Jen Chu and Sean Crane and Avijit Dasgupta and Jing Dong and Maria Escobar and Cristhian Forigua and Abrham Gebreselasie and Sanjay Haresh and Jing Huang and Md Mohaiminul Islam and Suyog Jain and Rawal Khirodkar and Devansh Kukreja and Kevin J Liang and Jia-Wei Liu and Sagnik Majumder and Yongsen Mao and Miguel Martin and Effrosyni Mavroudi and Tushar Nagarajan and Francesco Ragusa and Santhosh Kumar Ramakrishnan and Luigi Seminara and Arjun Somayazulu and Yale Song and Shan Su and Zihui Xue and Edward Zhang and Jinxu Zhang and Angela Castillo and Changan Chen and Xinzhu Fu and Ryosuke Furuta and Cristina Gonzalez and Prince Gupta and Jiabo Hu and Yifei Huang and Yiming Huang and Weslie Khoo and Anush Kumar and Robert Kuo and Sach Lakhavani and Miao Liu and Mi Luo and Zhengyi Luo and Brighid Meredith and Austin Miller and Oluwatumininu Oguntola and Xiaqing Pan and Penny Peng and Shraman Pramanick and Merey Ramazanova and Fiona Ryan and Wei Shan and Kiran Somasundaram and Chenan Song and Audrey Southerland and Masatoshi Tateno and Huiyu Wang and Yuchen Wang and Takuma Yagi and Mingfei Yan and Xitong Yang and Zecheng Yu and Shengxin Cindy Zha and Chen Zhao and Ziwei Zhao and Zhifan Zhu and Jeff Zhuo and Pablo Arbelaez and Gedas Bertasius and David Crandall and Dima Damen and Jakob Engel and Giovanni Maria Farinella and Antonino Furnari and Bernard Ghanem and Judy Hoffman and C. V. Jawahar and Richard Newcombe and Hyun Soo Park and James M. Rehg and Yoichi Sato and Manolis Savva and Jianbo Shi and Mike Zheng Shou and Michael Wray},
      year={2024},
    booktitle={Proceedings of the IEEE/CVF Conference on Computer Vision and Pattern Recognition},
    pages={19383-19400},
}

@inproceedings{hampali2020ho3d,
  title     = {{HOnnotate}: A Method for 3D Annotation of Hand and Object Poses},
  author    = {Hampali, Shreyas and Rad, Mahdi and Oberweger, Markus and Lepetit, Vincent},
  booktitle = {Proceedings of the IEEE/CVF Conference on Computer Vision and Pattern Recognition},
  year      = {2020}
}

@InProceedings{keypointtransformer2022,  
author = {Shreyas Hampali and Sayan Deb Sarkar and Mahdi Rad and Vincent Lepetit},  
title = {Keypoint Transformer: Solving Joint Identification in Challenging Hands and Object Interactions for Accurate 3D Pose Estimation},  
booktitle = {IEEE Computer Vision and Pattern Recognition Conference},  
year = {2022}  
}

@inproceedings{
    egodex2025,
    title={EgoDex: Learning Dexterous Manipulation from Large-Scale Egocentric Video},
    author={Ryan Hoque and Peide Huang and David J. Yoon and Mouli sivapurapu and Jian Zhang},
    booktitle={The Fourteenth International Conference on Learning Representations},
    year={2026},
    url={https://openresix.net/pdf/d4fea01944202c75aa13dc34757e154478eb8df7.pdf}
}

@misc{hand3r2026,
      title={Hand3R: Online 4D Hand-Scene Reconstruction in the Wild}, 
      author={Wendi Hu and Haonan Zhou and Wenhao Hu and Gaoang Wang},
      year={2026},
      eprint={2602.03200},
      archivePrefix={arXiv},
      primaryClass={cs.CV},
      url={https://arxiv.org/abs/2602.03200}, 
}

@misc{vipe2025,
      title={ViPE: Video Pose Engine for 3D Geometric Perception}, 
      author={Jiahui Huang and Qunjie Zhou and Hesam Rabeti and Aleksandr Korovko and Huan Ling and Xuanchi Ren and Tianchang Shen and Jun Gao and Dmitry Slepichev and Chen-Hsuan Lin and Jiawei Ren and Kevin Xie and Joydeep Biswas and Laura Leal-Taixe and Sanja Fidler},
      year={2025},
      eprint={2508.10934},
      archivePrefix={arXiv},
      primaryClass={cs.CV},
      url={https://arxiv.org/abs/2508.10934}, 
}

@inproceedings{
jing2025humanoidgen,
title={HumanoidGen: Data Generation for Bimanual Dexterous Manipulation via {LLM} Reasoning},
author={Zhi Jing and Siyuan Yang and Jicong Ao and Ting Xiao and Yu-Gang Jiang and Chenjia Bai},
booktitle={The Thirty-ninth Annual Conference on Neural Information Processing Systems},
year={2025},
url={https://openreview.net/forum?id=Mk9ykil8eP}
}

@inproceedings{mapanything2025,
  title={{MapAnything}: Universal Feed-Forward Metric {3D} Reconstruction},
  author={Nikhil Keetha and Norman M\"{u}ller and Johannes Sch\"{o}nberger and Lorenzo Porzi and Yuchen Zhang and Tobias Fischer and Arno Knapitsch and Duncan Zauss and Ethan Weber and Nelson Antunes and Jonathon Luiten and Manuel Lopez-Antequera and Samuel Rota Bul\`{o} and Christian Richardt and Deva Ramanan and Sebastian Scherer and Peter Kontschieder},
  booktitle={International Conference on 3D Vision (3DV)},
  year={2026},
  organization={IEEE}
}

@InProceedings{kwon2021h2o,
author = {Kwon, Taein and Tekin, Bugra and St\"uhmer, Jan and Bogo, Federica and Pollefeys, Marc},
title = {H2O: Two Hands Manipulating Objects for First Person Interaction Recognition},
booktitle = {Proceedings of the IEEE/CVF International Conference on Computer Vision (ICCV)},
month = {October},
year = {2021},
pages = {10138-10148}
}

@inproceedings{labbe2022megapose,
    title     = {MegaPose: 6D Pose Estimation of Novel Objects via Render \& Compare},
    author    = {Labb\'e, Yann and Manuelli, Lucas and Mousavian, Arsalan and Tyree, Stephen and Birchfield, Stan and Tremblay, Jonathan and Carpentier, Justin and Aubry, Mathieu and Fox, Dieter and Sivic, Josef},
    booktitle = {Proceedings of the 6th Conference on Robot Learning (CoRL)},
    year      = {2022},
  }

@inproceedings{any6d2025,
    title     = {{Any6D}: Model-free 6D Pose Estimation of Novel Objects},
    author    = {Lee, Taeyeop and Wen, Bowen and Kang, Minjun and Kang, Gyuree and Kweon, In So and Yoon, Kuk-Jin},
    booktitle = {Proceedings of the Computer Vision and Pattern Recognition Conference (CVPR)},
    year      = {2025},
}

@article{mvsam3d2026,
  title={MV-SAM3D: Adaptive Multi-View Fusion for Layout-Aware 3D Generation},
  author={Li, Baicheng and Wu, Dong and Li, Jun and Zhou, Shunkai and Zeng, Zecui and Li, Lusong and Zha, Hongbin},
  journal={arXiv preprint arXiv:2603.11633},
  year={2026}
}

@inproceedings{
iggt2026,
title={{IGGT}: Instance-Grounded Geometry Transformer for Semantic 3D Reconstruction},
author={Hao Li and Zhengyu Zou and Fangfu Liu and Xuanyang Zhang and Fangzhou Hong and Yukang Cao and Yushi LAN and Manyuan Zhang and Gang YU and Dingwen Zhang and Ziwei Liu},
booktitle={The Fourteenth International Conference on Learning Representations},
year={2026},
url={https://openreview.net/forum?id=swiL18PmUV}
}

@inproceedings{megasam2025,
  title     = {{MegaSaM: Accurate, Fast and Robust Structure and Motion from Casual Dynamic Videos}},
  author    = {Li, Zhengqi and Tucker, Richard and Cole, Forrester and Wang, Qianqian and Jin, Linyi and Ye, Vickie and Kanazawa, Angjoo and Holynski, Aleksander and Snavely, Noah},
  booktitle = {Proceedings of the IEEE/CVF Conference on Computer Vision and Pattern Recognition},
  year      = {2025}
}

@article{hggt2026,
      title={HGGT: Robust and Flexible 3D Hand Mesh Reconstruction from Uncalibrated Images},
      author={Liu, Yumeng and Long, Xiao-Xiao and Habermann, Marc and Yang, Xuanze and Lin, Cheng and Liu, Yuan and Ma, Yuexin and Liu, Ligang},
      journal={arXiv preprint arXiv:2603.23997},
      year={2026}
    }

@inproceedings{liu2024taco,
  title = {{TACO: Benchmarking Generalizable Bimanual Tool-ACtion-Object Understanding}},
  author = {Yun Liu and Haolin Yang and Xu Si and Ling Liu and Zipeng Li and Yuxiang Zhang and Yebin Liu and Li Yi},
  booktitle = {Proceedings of the IEEE/CVF Conference on Computer Vision and Pattern Recognition},
  year = {2024}
}

@InProceedings{liu2022hoi4d,
    author    = {Liu, Yunze and Liu, Yun and Jiang, Che and Lyu, Kangbo and Wan, Weikang and Shen, Hao and Liang, Boqiang and Fu, Zhoujie and Wang, He and Yi, Li},
    title     = {HOI4D: A 4D Egocentric Dataset for Category-Level Human-Object Interaction},
    booktitle = {Proceedings of the IEEE/CVF Conference on Computer Vision and Pattern Recognition (CVPR)},
    month     = {June},
    year      = {2022},
    pages     = {21013-21022}
}

@inproceedings{reviv2026,
  title     = {ReViV: Reconstructing the Viewer and the View in 4D from Monocular Egocentric Video},
  author    = {Lyu, Xiaozhong and Li, Gen and Qian, Zhiyin and Zhang, Xucong and Pollefeys, Marc and Tang, Siyu},
  booktitle = {European Conference on Computer Vision (ECCV)},
  year      = {2026}
}

@inproceedings{olson2011apriltag,
  title={AprilTag: A robust and flexible visual fiducial system},
  author={Olson, Edwin},
  booktitle={Proceedings of the {IEEE} International Conference on Robotics and Automation ({ICRA})},
  pages={3400--3407},
  year={2011},
}

@misc{adt2023,
      title={Aria Digital Twin: A New Benchmark Dataset for Egocentric 3D Machine Perception}, 
      author={Xiaqing Pan and Nicholas Charron and Yongqian Yang and Scott Peters and Thomas Whelan and Chen Kong and Omkar Parkhi and Richard Newcombe and Carl Yuheng Ren},
      year={2023},
      eprint={2306.06362},
      archivePrefix={arXiv},
      primaryClass={cs.CV},
      url={https://arxiv.org/abs/2306.06362}, 
}

@inproceedings{pavlakos2024hamer,
    title={Reconstructing Hands in 3D with Transformers},
    author={Pavlakos, Georgios and Shan, Dandan and Radosavovic, Ilija and Kanazawa, Angjoo and Fouhey, David and Malik, Jitendra},
    booktitle={Proceedings of the IEEE/CVF Conference on Computer Vision and Pattern Recognition (CVPR)},
    year={2024}
}

@InProceedings{hdepic2025,
  author    = {Perrett, Toby and Darkhalil, Ahmad and Sinha, Saptarshi and Emara, Omar and Pollard, Sam and Parida, Kranti and Liu, Kaiting and Gatti, Prajwal and Bansal, Siddhant and Flanagan, Kevin and Chalk, Jacob and Zhu, Zhifan and Guerrier, Rhodri and Abdelazim, Fahd and Zhu, Bin and Moltisanti, Davide and Wray, Michael and Doughty, Hazel and Damen, Dima},
  title     = {HD-EPIC: A Highly-Detailed Egocentric Video Dataset},
  booktitle = {Proceedings of the IEEE/CVF Conference on Computer Vision and Pattern Recognition (CVPR)},
  year      = {2025},
  month     = {June}
}

@inproceedings{unidepth2024,
    title     = {UniDepth: Universal Monocular Metric Depth Estimation},
    author    = {Piccinelli, Luigi and Yang, Yung-Hsu and Sakaridis, Christos and Segu, Mattia and Li, Siyuan and Van Gool, Luc and Yu, Fisher},
    booktitle = {Proceedings of the IEEE/CVF Conference on Computer Vision and Pattern Recognition (CVPR)},
    year      = {2024}
}

@inproceedings{potamias2025wilor,
  title     = {{WiLoR}: End-to-end 3D Hand Localization and Reconstruction in-the-wild},
  author    = {Potamias, Rolandos Alexandros and Zhang, Jinglei and Deng, Jiankang and Zafeiriou, Stefanos},
  booktitle = {Proceedings of the IEEE/CVF Conference on Computer Vision and Pattern Recognition (CVPR)},
  year      = {2025}
}

@misc{egoverse2026,
      title={EgoVerse: An Egocentric Human Dataset for Robot Learning from Around the World}, 
      author={Ryan Punamiya and Simar Kareer and Zeyi Liu and Josh Citron and Ri-Zhao Qiu and Xiongyi Cai and Alexey Gavryushin and Jiaqi Chen and Davide Liconti and Lawrence Y. Zhu and Patcharapong Aphiwetsa and Baoyu Li and Aniketh Cheluva and Pranav Kuppili and Yangcen Liu and Dhruv Patel and Aidan Gao and Hye-Young Chung and Ryan Co and Renee Zbizika and Jeff Liu and Xiaomeng Xu and Haoyu Xiong and Geng Chen and Sebastiano Oliani and Wenkai Xuan and Chenyu Yang and Xi Wang and James Fort and Richard Newcombe and Josh Gao and Jason Chong and Garrett Matsuda and Aseem Doriwala and Marc Pollefeys and Robert Katzschmann and Xiaolong Wang and Shuran Song and Judy Hoffman and Danfei Xu},
      year={2026},
      eprint={2604.07607},
      archivePrefix={arXiv},
      primaryClass={cs.RO},
      url={https://arxiv.org/abs/2604.07607}, 
}

@inproceedings{show3d2026,
  author    = {Patrick Rim and Kevin Harris and Braden Copple and Shangchen Han and Xu Xie and Ivan Shugurov and Sizhe An and He Wen and Alex Wong and Tomas Hodan and Kun He},
  title     = {{SHOW3D}: Capturing Scenes of 3D Hands and Objects in the Wild},
  booktitle = {Proceedings of the IEEE/CVF Conference on Computer Vision and Pattern Recognition (CVPR)},
  year      = {2026},
}

@article{romero2017mano,
      title = {Embodied Hands: Modeling and Capturing Hands and Bodies Together},
      author = {Romero, Javier and Tzionas, Dimitrios and Black, Michael J.},
      journal = {ACM Transactions on Graphics, (Proc. SIGGRAPH Asia)},
      volume = {36},
      number = {6},
      series = {245:1--245:17},
      month = nov,
      year = {2017},
      month_numeric = {11}
  }

@misc{dinov3_2025,
  title={{DINOv3}},
  author={Sim{\'e}oni, Oriane and Vo, Huy V. and Seitzer, Maximilian and Baldassarre, Federico and Oquab, Maxime and Jose, Cijo and Khalidov, Vasil and Szafraniec, Marc and Yi, Seungeun and Ramamonjisoa, Micha{\"e}l and Massa, Francisco and Haziza, Daniel and Wehrstedt, Luca and Wang, Jianyuan and Darcet, Timoth{\'e}e and Moutakanni, Th{\'e}o and Sentana, Leonel and Roberts, Claire and Vedaldi, Andrea and Tolan, Jamie and Brandt, John and Couprie, Camille and Mairal, Julien and J{\'e}gou, Herv{\'e} and Labatut, Patrick and Bojanowski, Piotr},
  year={2025},
  eprint={2508.10104},
  archivePrefix={arXiv},
  primaryClass={cs.CV},
  url={https://arxiv.org/abs/2508.10104},
}

@misc{dpm2025,
      title={Dynamic Point Maps: A Versatile Representation for Dynamic 3D Reconstruction}, 
      author={Edgar Sucar and Zihang Lai and Eldar Insafutdinov and Andrea Vedaldi},
      year={2025},
      eprint={2503.16318},
      archivePrefix={arXiv},
      primaryClass={cs.CV},
      url={https://arxiv.org/abs/2503.16318}, 
}

@inproceedings{vdpm2026,
  title={{V-DPM}: 4D Video Reconstruction with Dynamic Point Maps},
  author={Sucar, Edgar and Insafutdinov, Eldar and Lai, Zihang and Vedaldi, Andrea},
  booktitle={Proceedings of the IEEE/CVF Conference on Computer Vision and Pattern Recognition (CVPR)},
  year={2026}
}

@inproceedings{wang2025vggt,
  title={VGGT: Visual Geometry Grounded Transformer},
  author={Wang, Jianyuan and Chen, Minghao and Karaev, Nikita and Vedaldi, Andrea and Rupprecht, Christian and Novotny, David},
  booktitle={Proceedings of the IEEE/CVF Conference on Computer Vision and Pattern Recognition},
  year={2025}
}

@inproceedings{vggtomega2026,
  title     = {{VGGT-$\Omega$}},
  author    = {Jianyuan Wang and Minghao Chen and Shangzhan Zhang and Nikita Karaev and Johannes Sch{\"o}nberger and Patrick Labatut and Piotr Bojanowski and David Novotny and Andrea Vedaldi and Christian Rupprecht},
  booktitle = {Proceedings of the IEEE/CVF Conference on Computer Vision and Pattern Recognition (CVPR)},
  year      = {2026}
}

@inproceedings{hocap2024,
title={{HO}-Cap: A Capture System and Dataset for 3D Reconstruction and Pose Tracking of Hand-Object Interaction},
author={Jikai Wang and Qifan Zhang and Yu-Wei Chao and Bowen Wen and Xiaohu Guo and Yu Xiang},
booktitle={The Thirty-ninth Annual Conference on Neural Information Processing Systems Datasets and Benchmarks Track},
year={2025},
url={https://openreview.net/forum?id=hpu6r8oLw9}
}

@inproceedings{wang2025cut3r,
    Author = {Qianqian Wang and Yifei Zhang and Aleksander Holynski and Alexei A. Efros and Angjoo Kanazawa},
    Title = {Continuous 3D Perception Model with Persistent State},
    Year = {2025},
      booktitle={Proceedings of the IEEE/CVF Conference on Computer Vision and Pattern Recognition (CVPR)},
    }

@inproceedings{wang2024dust3r,
      title={DUSt3R: Geometric 3D Vision Made Easy}, 
      author={Shuzhe Wang and Vincent Leroy and Yohann Cabon and Boris Chidlovskii and Jerome Revaud},
      booktitle = {CVPR},
      year = {2024}
}

@inproceedings{arthoi2026,
  title={ArtHOI: Taming Foundation Models for Monocular 4D Reconstruction of Hand-Articulated-Object Interactions},
  author={Wang, Zikai and Zhang, Zhilu and Wang, Yiqing and Li, Hui and Zuo, Wangmeng},
  booktitle={Proceedings of the IEEE/CVF Conference on Computer Vision and Pattern Recognition},
  year={2026}
}

@InProceedings{wen2024foundationpose,
author = {Bowen Wen and Wei Yang and Jan Kautz and Stan Birchfield},
title         = {{FoundationPose}: Unified 6D Pose Estimation and Tracking of Novel Objects},
booktitle     = {Proceedings of the IEEE/CVF Conference on Computer Vision and Pattern Recognition},
year          = {2024},
}

@article{whole2026,
    author = {Ye, Yufei and Li, Jiaman and Rong, Ryan and Liu, C. Karen},
    title = {WHOLE: World-Grounded Hand-Object Lifted from Egocentric Videos},
    journal = {CVPR Findings},
    year = {2026}
}

@inproceedings{yin2023metric3d,
  title     = {{Metric3D}: Towards Zero-shot Metric {3D} Prediction from a Single Image},
  author    = {Yin, Wei and
               Zhang, Chi and
               Chen, Hao and
               Cai, Zhipeng and
               Yu, Gang and
               Wang, Kaixuan and
               Chen, Xiaozhi and
               Shen, Chunhua},
  booktitle = {Proceedings of the IEEE/CVF International Conference on Computer Vision},
  year      = {2023}
}

@misc{yin2025dexteritygen,
      title={DexterityGen: Foundation Controller for Unprecedented Dexterity}, 
      author={Zhao-Heng Yin and Changhao Wang and Luis Pineda and Francois Hogan and Krishna Bodduluri and Akash Sharma and Patrick Lancaster and Ishita Prasad and Mrinal Kalakrishnan and Jitendra Malik and Mike Lambeta and Tingfan Wu and Pieter Abbeel and Mustafa Mukadam},
      year={2025},
      eprint={2502.04307},
      archivePrefix={arXiv},
      primaryClass={cs.RO},
      url={https://arxiv.org/abs/2502.04307}, 
}

@inproceedings{dynhamr2025,
  title={Dyn-HaMR: Recovering 4D Interacting Hand Motion from a Dynamic Camera},
  author={Yu, Zhengdi and Zafeiriou, Stefanos and Birdal, Tolga},
  booktitle = {Proceedings of the IEEE/CVF Conference on Computer Vision and Pattern Recognition (CVPR)},
  month     = {June},
  year      = {2025},
}

@inproceedings{d4rt2026,
  title={Efficiently Reconstructing Dynamic Scenes One D4RT at a Time},
  author={Zhang, Chuhan and Le Moing, Guillaume and Koppula, Skanda and Rocco, Ignacio and Momeni, Liliane and Xie, Junyu and Sun, Shuyang and Sukthankar, Rahul and Barral, Jo{\"e}lle K. and Hadsell, Raia and Ghahramani, Zoubin and Zisserman, Andrew and Zhang, Junlin and Sajjadi, Mehdi S. M.},
  booktitle={CVPR},
  year={2026}
}

@inproceedings{hawor2025,
      title={HaWoR: World-Space Hand Motion Reconstruction from Egocentric Videos},
      author={Zhang, Jinglei and Deng, Jiankang and Ma, Chao and Potamias, Rolandos Alexandros},
      booktitle={Proceedings of the IEEE/CVF Conference on Computer Vision and Pattern Recognition (CVPR)},
      year={2025}
    }

@inproceedings{
zhang2025monst3r,
title={Mon{ST}3R: A Simple Approach for Estimating Geometry in the Presence of Motion},
author={Junyi Zhang and Charles Herrmann and Junhwa Hur and Varun Jampani and Trevor Darrell and Forrester Cole and Deqing Sun and Ming-Hsuan Yang},
booktitle={The Thirteenth International Conference on Learning Representations},
year={2025},
url={https://openreview.net/forum?id=lJpqxFgWCM}
}

@misc{mint2026,
      title={MINT: A Unified Model for World-Space Camera and Hand Motion Estimation from Scalable Egocentric Pipeline Supervision}, 
      author={Zijie Zhu and Weiren Cai and Yizhou Wang and Zhenjie Yang and Yide Liu and Jiahao Chen and Guanqi He},
      year={2026},
      eprint={2609.04958},
      archivePrefix={arXiv},
      primaryClass={cs.CV},
      url={https://arxiv.org/abs/2609.04958}, 
}

@article{da3_2025,
  title   = {{Depth Anything 3: Recovering the Visual Space from Any Views}},
  author  = {Haotong Lin and Sili Chen and Junhao Liew and Donny Y. Chen and Zhenyu Li and Guang Shi and Jiashi Feng and Bingyi Kang},
  journal = {arXiv preprint arXiv:2511.10647v1},
  year    = {2025},
  note    = {\url{https://depth-anything-3.github.io/}},
}

@inproceedings{gigapose2024,
  title = {{GigaPose: Fast and Robust Novel Object Pose Estimation via One Correspondence}},
  author = {Van Nguyen Nguyen and Thibault Groueix and Mathieu Salzmann and Vincent Lepetit},
  booktitle = {Proceedings of the IEEE/CVF Conference on Computer Vision and Pattern Recognition (CVPR)},
  year = {2024},
  url = {https://openaccess.thecvf.com/content/CVPR2024/html/Nguyen_GigaPose_Fast_and_Robust_Novel_Object_Pose_Estimation_via_One_CVPR_2024_paper.html}
}

@misc{homie2026,
  title        = {{HOMIE Gen2: Scale Real-World Data for Physical AI}},
  author       = {{Ropedia R\&D Team}},
  year         = {2026},
  howpublished = {\url{https://ropedia.com/homie}},
  note         = {Accessed 2026-09-19}
}

@article{handflow2026,
  title = {{HandFlow: Fully Generative 4D Hand Recovery with Flow Matching}},
  author = {Xu, Mingxi and Duan, Bowen and Gu, Yi and Shen, Zhengyang and Xu, Renjing and Yue, Yutao},
  journal = {arXiv preprint arXiv:2607.11221},
  year = {2026}
}

@inproceedings{cao2026hsimul3r,
  title={HSImul3R: Physics-in-the-Loop Reconstruction of Simulation-Ready Human--Scene Interactions},
  author={Cao, Yukang and Xie, Haozhe and Hong, Fangzhou and Zhuo, Long and Chen, Zhaoxi and Pan, Liang and Liu, Ziwei},
  booktitle={European Conference on Computer Vision},
  pages={509--528},
  year={2026},
  organization={Springer}
}

@article{cao2026crowdmogen,
  title={Crowdmogen: Event-driven collective human motion generation},
  author={Cao, Yukang and Guo, Xinying and Zhang, Mingyuan and Xie, Haozhe and Gu, Chenyang and Liu, Ziwei},
  journal={International Journal of Computer Vision},
  volume={134},
  number={1},
  pages={29},
  year={2026},
  publisher={Springer}
}

@misc{xperience_10m,
  title = {{Xperience-10M}: A Large-Scale Egocentric Multimodal Dataset with Structured {3D/4D} Annotations},
  author = {{Ropedia}},
  year = {2026},
  howpublished = {Hugging Face},
  url = {https://huggingface.co/datasets/ropedia-ai/xperience-10m},
  note = {Dataset}
}

@misc{agibotworld2026,
    title        = {AgiBot World 2026},
    author       = {{AgiBot World Team}},
    howpublished = {\url{https://huggingface.co/datasets/agibot-world/AgiBotWorld2026}},
    year         = {2026}
}

@misc{avetisyan2024scenescriptreconstructingscenesautoregressive,
      title={SceneScript: Reconstructing Scenes With An Autoregressive Structured Language Model}, 
      author={Armen Avetisyan and Christopher Xie and Henry Howard-Jenkins and Tsun-Yi Yang and Samir Aroudj and Suvam Patra and Fuyang Zhang and Duncan Frost and Luke Holland and Campbell Orme and Jakob Engel and Edward Miller and Richard Newcombe and Vasileios Balntas},
      year={2024},
      eprint={2403.13064},
      archivePrefix={arXiv},
      primaryClass={cs.CV},
      url={https://arxiv.org/abs/2403.13064}, 
}

@article{li2024behavior1k,
    title   = {BEHAVIOR-1K: A Human-Centered, Embodied AI Benchmark with 1,000 Everyday Activities and Realistic Simulation},
    author  = {Chengshu Li and Ruohan Zhang and Josiah Wong and Cem Gokmen and Sanjana Srivastava and Roberto Martín-Martín and Chen Wang and Gabrael Levine and Wensi Ai and Benjamin Martinez and Hang Yin and Michael Lingelbach and Minjune Hwang and Ayano Hiranaka and Sujay Garlanka and Arman Aydin and Sharon Lee and Jiankai Sun and Mona Anvari and Manasi Sharma and Dhruva Bansal and Samuel Hunter and Kyu-Young Kim and Alan Lou and Caleb R Matthews and Ivan Villa-Renteria and Jerry Huayang Tang and Claire Tang and Fei Xia and Yunzhu Li and Silvio Savarese and Hyowon Gweon and C. Karen Liu and Jiajun Wu and Li Fei-Fei},
    journal = {arXiv preprint arXiv:2403.09227},
    year    = {2024}
}

@InProceedings{McCormac_2017_ICCV,
author = {McCormac, John and Handa, Ankur and Leutenegger, Stefan and Davison, Andrew J.},
title = {SceneNet RGB-D: Can 5M Synthetic Images Beat Generic ImageNet Pre-Training on Indoor Segmentation?},
booktitle = {Proceedings of the IEEE International Conference on Computer Vision (ICCV)},
month = {Oct},
year = {2017}
}

@article{karaev2023dynamicstereo,
  title={DynamicStereo: Consistent Dynamic Depth from Stereo Videos},
  author={Nikita Karaev and Ignacio Rocco and Benjamin Graham and Natalia Neverova and Andrea Vedaldi and Christian Rupprecht},
  journal={CVPR},
  year={2023}
}
}

\newpage
\appendix
\startcontents[appendix]
\printcontents[appendix]{l}{1}{%
  \section*{Appendix Contents}%
  \vspace{-0.5em}%
}
\newpage
\section{Extended Related Work}
\label{app:related}

\subsection{Joint geometry and entity reconstruction}

DUSt3R~\citep{wang2024dust3r} predicts pairwise pointmaps, while
VGGT~\citep{wang2025vggt} jointly estimates cameras and geometry.
CUT3R~\citep{wang2025cut3r} uses recurrent scene memory, and
VGGT-$\Omega$~\citep{vggtomega2026} reduces inter-frame attention
costs with scene registers. EgoX combines register-mediated and
global attention with hand, object, and scale tokens. For dynamic
scenes, MonST3R~\citep{zhang2025monst3r} reconstructs moving content,
and St4RTrack~\citep{st4rtrack2025} tracks 3D points in a common world
frame. Dynamic Point Maps~\citep{dpm2025} recovers rigid-object motion
through masked point alignment, V-DPM~\citep{vdpm2026} extends
pointmaps to video, and D4RT~\citep{d4rt2026} predicts 3D points
conditioned on source, target, and camera-reference times. EgoX
directly predicts articulated hands and rigid object poses alongside
cameras and depth. Human3R~\citep{human3r2025} combines human meshes
with scene and camera reconstruction, while IGGT~\citep{iggt2026}
jointly learns geometry and instance segmentation. HGGT~\citep{hggt2026}
uses hand tokens to estimate MANO parameters and cameras from
uncalibrated views of a fixed hand configuration. Unlike these methods,
EgoX jointly reconstructs the camera, depth, and moving hands and
objects over time.

\subsection{Hand and hand--object reconstruction}

HaMeR~\citep{pavlakos2024hamer} regresses MANO~\citep{romero2017mano}
parameters from hand crops, while WiLoR~\citep{potamias2025wilor}
pairs hand detection with mesh estimation. For world-space motion,
HaWoR~\citep{hawor2025} combines hand reconstruction with SLAM and
motion infilling; Dyn-HaMR~\citep{dynhamr2025} uses camera tracking,
motion priors, and optimization. Hand3R~\citep{hand3r2026} couples a
hand estimator with recurrent scene modeling, while MINT~\citep{mint2026}
and ReViV~\citep{reviv2026} jointly estimate hand and camera motion,
with ReViV also modeling depth, body motion, and gaze. EgoX additionally
predicts manipulated-object motion.

Simulation-ready objects~\citep{physx,physxanything,physxomni} and
synthetic motion~\citep{jing2025humanoidgen,yin2025dexteritygen}
enable interaction data at scale, but may not capture natural human
manipulation. Keypoint Transformer~\citep{keypointtransformer2022}
and HOPformer~\citep{hopformer2026} estimate hand--object interactions
without a world-camera trajectory. HOLD~\citep{fan2024hold} optimizes
hand and object representations from video; ForeHOI~\citep{forehoi2026}
completes masks and shapes before geometrically estimating object poses,
without requiring separate input masks. ArtHOI~\citep{arthoi2026}
handles articulated objects, whereas EgoX models rigid objects.

For world-space interactions, EgoGrasp~\citep{egograsp2026}
reconstructs hands and objects in separate stages. WHOLE~\citep{whole2026}
uses a motion prior and test-time guidance but requires object templates
and a metric camera trajectory. EgoX jointly predicts cameras, depth,
hands, and objects from RGB video and initial object masks, without
either external input. Its reconstruction losses evaluate hand, object,
and depth predictions through the predicted camera, providing training
signals for camera estimation.

\subsection{Camera, depth, and object-pose pipelines}

Modular reconstruction also provides a source of training annotations.
MegaSaM~\citep{megasam2025} and ViPE~\citep{vipe2025} combine learned
geometry with camera and depth optimization. Metric3D~\citep{yin2023metric3d}
and UniDepth~\citep{unidepth2024} predict metric depth from single
images, while MapAnything~\citep{mapanything2025} predicts metric
geometry from multiple views. HaWoR uses metric depth to scale its
SLAM trajectory. EgoX instead learns a scale per window shared by
depth and camera, hand, and object translations, though shared scale
alone does not ensure geometric consistency.

MegaPose~\citep{labbe2022megapose} requires a CAD model,
FoundationPose~\citep{wen2024foundationpose} accepts a model or
reference images, and Any6D~\citep{any6d2025} uses an RGB-D reference.
Their camera-relative poses can be mapped to world coordinates using
a camera trajectory. EgoX likewise transforms predictions into a
common frame, but learns camera and object estimates jointly. These
methods also differ in their requirements for depth, pose
initialization, scale, and masks.

\subsection{Datasets and geometric evaluation}

HO-3D~\citep{hampali2020ho3d}, DexYCB~\citep{chao2021dexycb},
H2O~\citep{kwon2021h2o}, H2O-3D~\citep{keypointtransformer2022},
HOI4D~\citep{liu2022hoi4d}, ARCTIC~\citep{fan2023arctic},
TACO~\citep{liu2024taco}, HO-Cap~\citep{hocap2024}, and
HOT3D~\citep{hot3d2024} support hand--object reconstruction under
varied capture protocols. ARCTIC includes articulated objects, and
HOT3D provides motion-capture reference poses. However, capture
volumes or clip protocols often limit evaluation of continuous camera
travel between workspaces.

More recent datasets broaden the settings. SHOW3D~\citep{show3d2026}
captures indoor and outdoor interactions with a mobile ego--exo rig,
but defines annotations relative to the moving rig.
EPIC-Contact~\citep{hopformer2026} provides fitted meshes and contact
correspondences in kitchen video. Aria Digital Twin~\citep{adt2023}
provides camera and object trajectories, scanned scenes, and rendered
depth; HD-EPIC~\citep{hdepic2025} provides unscripted kitchen
recordings with digital twins and 3D object motion, though its labels
serve a different purpose from dense hand and object pose trajectories.

Large collections such as Ego4D~\citep{grauman2022ego4d} and
EPIC-KITCHENS~\citep{damen2022epickitchens} support learned geometric
supervision. EgoDex~\citep{egodex2025} scales manipulation data, while
MINT~\citep{mint2026} builds a camera-and-hand corpus through automatic
annotation. Our data spans workspace manipulation, travel between
indoor workspaces, and outdoor locomotion, with hand, object, and scene
annotations. This enables evaluation across camera displacement,
capture setting, and sequence duration using camera, depth, and
world-space interaction errors.



\section{Public Dataset Cleaning and Geometric Alignment}
\label{app:depth}

We systematically clean HOT3D, HOI4D, H2O, H2O-3D, HO-Cap,
and DexYCB before joint training. We discard frames with erroneous
hand annotations and remove spurious points from depth-derived
point clouds. We further align depth-derived geometry, MANO hand
meshes, and object meshes in a common 3D coordinate system,
correcting inconsistencies in their relative placement and scale.
These steps reduce erroneous labels and conflicting geometric
supervision across reconstruction tasks.

\section{Evaluation Protocol}
\label{app:evaluation}
\label{sec:setup}

\paragraph{Datasets and splits.}
We evaluate on four sequence-disjoint benchmarks: HaWoR's validation
split of HOT3D~\citep{hawor2025}, the held-out one-hour
EgoX-Bench, the official H2O test split, and our sequence-level
HOI4D validation split. Evaluation sequences are excluded from EgoX
training. All methods use identical sequence lists and matched
150-observation clips.

\paragraph{Inputs and matched protocol.}
EgoX receives RGB and optional first-frame object masks.
Object-pose baselines additionally use DA3 Nested depth and intrinsics,
per-frame masks, or metric CAD assets as required; their runtimes
include any required DA3 preprocessing. Any6D receives DA3 Nested depth
and intrinsics together with per-frame object masks, but no CAD asset.
FoundationPose receives DA3 Nested depth and intrinsics together with
the metric CAD asset. GigaPose + MegaPose receives DA3 Nested
intrinsics, the metric CAD asset, and per-frame object masks.
Ground-truth camera poses and scale are not supplied as reconstruction
inputs in the main comparisons.

For HandFlow, we follow the released right-hand-only protocol by
mirroring left-hand inputs and mapping recovered joints back to the
original camera frame. HandFlow and WiLoR are paired with the same
DROID-SLAM camera trajectory; neither receives ground-truth camera
poses, depth, or hand annotations. Results are averaged over matched
clips after applying common annotation-validity masks.

\paragraph{Metrics and coordinates.}
Hands are evaluated using PA-, W-, and WA-MPJPE. Cameras use normalized
absolute trajectory error (nATE) and relative rotation error (RRE);
depth uses absolute relative error (AbsRel) and threshold accuracy
$\delta_{1.25}$; object poses use ADD, ADD-S, and AUC-S$_{50}$.
World-space hand metrics use each pipeline's estimated camera
trajectory, whereas object pose is evaluated in camera coordinates.

\paragraph{Scale.}
EgoX and the native metric baselines retain their estimated metric
scale. VGGT and VGGT-$\Omega$ use one scale coefficient fitted from
DA3 Nested depth per input sequence. Alignment, aggregation, validity
masks, and failure handling are held fixed across methods.

\section{Runtime Analysis}
\label{app:runtime}

We measure method-level inference time for 150 frames on one NVIDIA
A800 (Table~\ref{tab:runtime}). Models and reusable assets are preloaded; data loading,
visualization, and disk I/O are excluded. Runtime for object-pose
baselines includes DA3 Nested whenever estimated depth or intrinsics
are required.

\definecolor{EGGTResultBlue}{HTML}{E2F3E8}
\begin{table}[H]
\centering
\caption{\textbf{Inference Efficiency.}
Method-level runtime for 150 frames on one NVIDIA A800.
Object baselines include DA3 Nested preprocessing when required.}
\label{tab:runtime}
\scriptsize
\setlength{\tabcolsep}{5pt}
\renewcommand{\arraystretch}{1.00}

\begin{adjustbox}{max width=\linewidth}
\begin{tabular}{@{}lcc@{\qquad}lcc@{}}
\toprule
\multicolumn{3}{c}{\textit{Hand reconstruction}} &
\multicolumn{3}{c}{\textit{Object 6D pose}} \\
\cmidrule(lr){1-3}
\cmidrule(lr){4-6}
Method & Time $\downarrow$ & FPS $\uparrow$
& Method & Time $\downarrow$ & FPS $\uparrow$ \\
\midrule
HaWoR & 25.37 s & 5.91
& Any6D & 166.54 s & 0.90 \\

MINT
& \cellcolor{EGGTResultBlue}\textbf{8.30 s}
& \cellcolor{EGGTResultBlue}\textbf{18.08}
& FoundationPose
& \underline{31.72 s}
& \underline{4.73} \\

ReViV & 14.91 s & 10.06
& GigaPose + MegaPose & 134.03 s & 1.12 \\

\textbf{EgoX}
& \underline{14.60 s}
& \underline{10.28}
& \textbf{EgoX}
& \cellcolor{EGGTResultBlue}\textbf{14.60 s}
& \cellcolor{EGGTResultBlue}\textbf{10.28} \\
\bottomrule
\end{tabular}
\end{adjustbox}
\end{table}

\section{Additional Qualitative Results}
\label{app:qualitative}

Figures~\ref{fig:hand_camera_comparison_additional},
\ref{fig:object_pose_comparison_additional}, and
\ref{fig:depth_comparison_additional} extend the main qualitative
comparisons with additional HOI4D and H2O sequences for world-space
hand and camera reconstruction, camera-frame object pose, and depth,
respectively.

\subsection{Additional hand and camera comparisons}
\label{app:additional_hand_camera}
\begin{figure}[H]
\centering
\includegraphics[width=\textwidth]{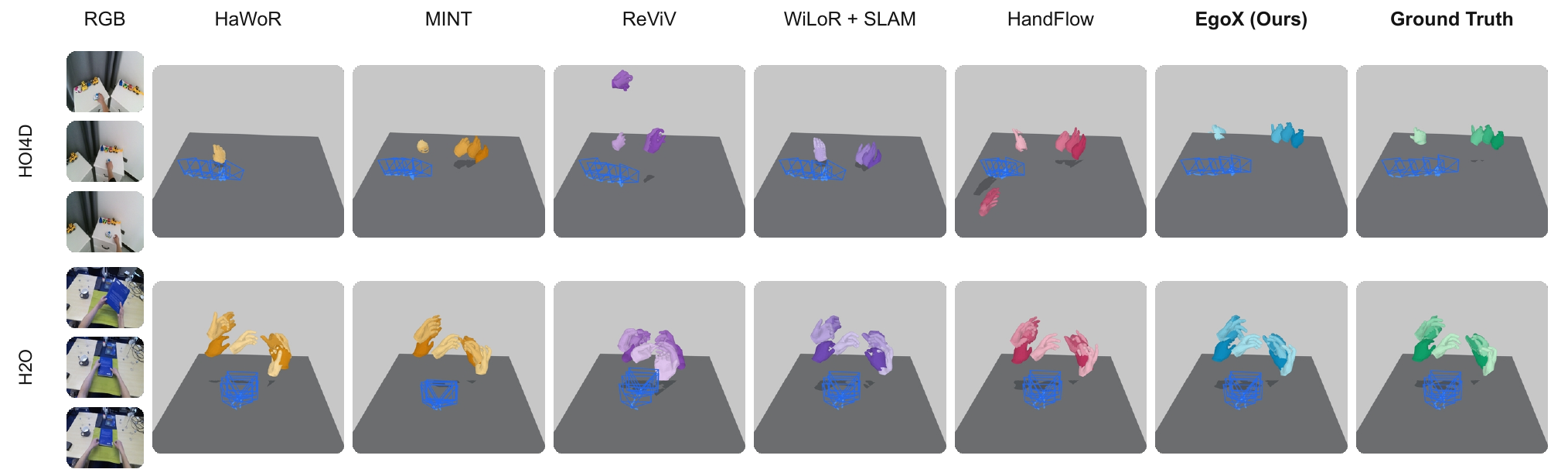}
\caption{\textbf{Additional World-Space Hand and Camera Reconstruction.}
Qualitative comparisons on HOI4D and H2O,
supplementing Figure~\ref{fig:hand_camera_comparison}.}
\label{fig:hand_camera_comparison_additional}
\end{figure}

Figure~\ref{fig:hand_camera_comparison_additional} supplements
Figure~\ref{fig:hand_camera_comparison} with additional world-space
hand and camera reconstructions.

\subsection{Additional object-pose comparisons}
\label{app:additional_object_pose}
\begin{figure}[H]
\centering
\includegraphics[width=\textwidth]{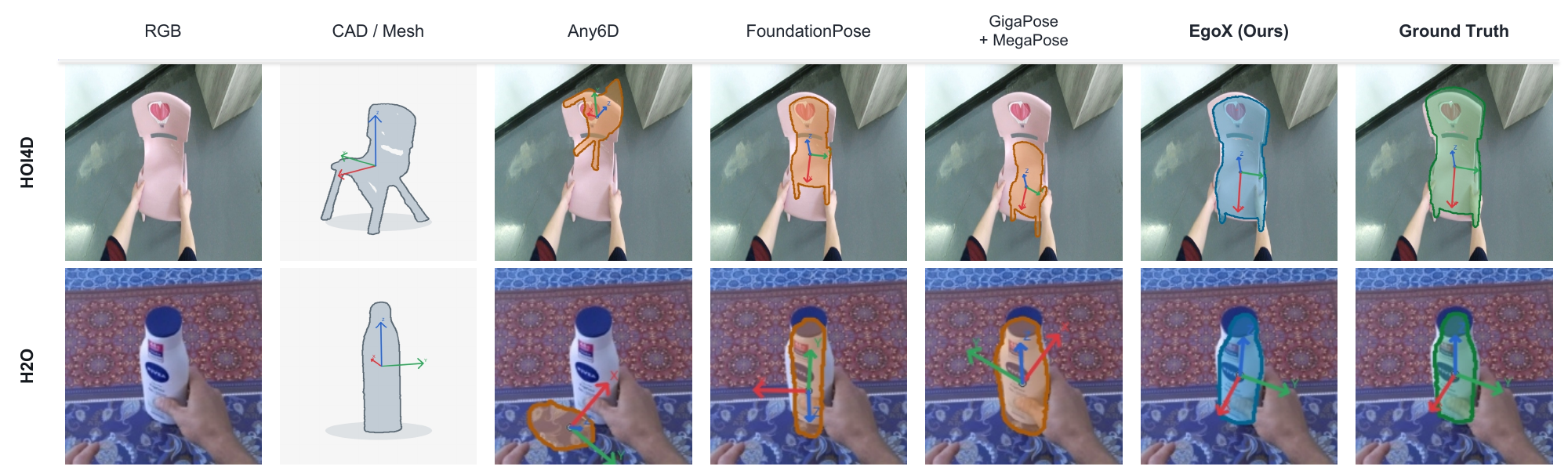}
\caption{\textbf{Additional Camera-Frame Object Pose Reconstruction.}
Qualitative comparisons on HOI4D and H2O,
supplementing Figure~\ref{fig:object_pose_comparison}.}
\label{fig:object_pose_comparison_additional}
\end{figure}

Figure~\ref{fig:object_pose_comparison_additional} supplements
Figure~\ref{fig:object_pose_comparison} with additional camera-frame
object-pose estimates.

\subsection{Additional depth comparisons}
\label{app:additional_depth}
\begin{figure}[H]
\centering
\includegraphics[width=\textwidth]{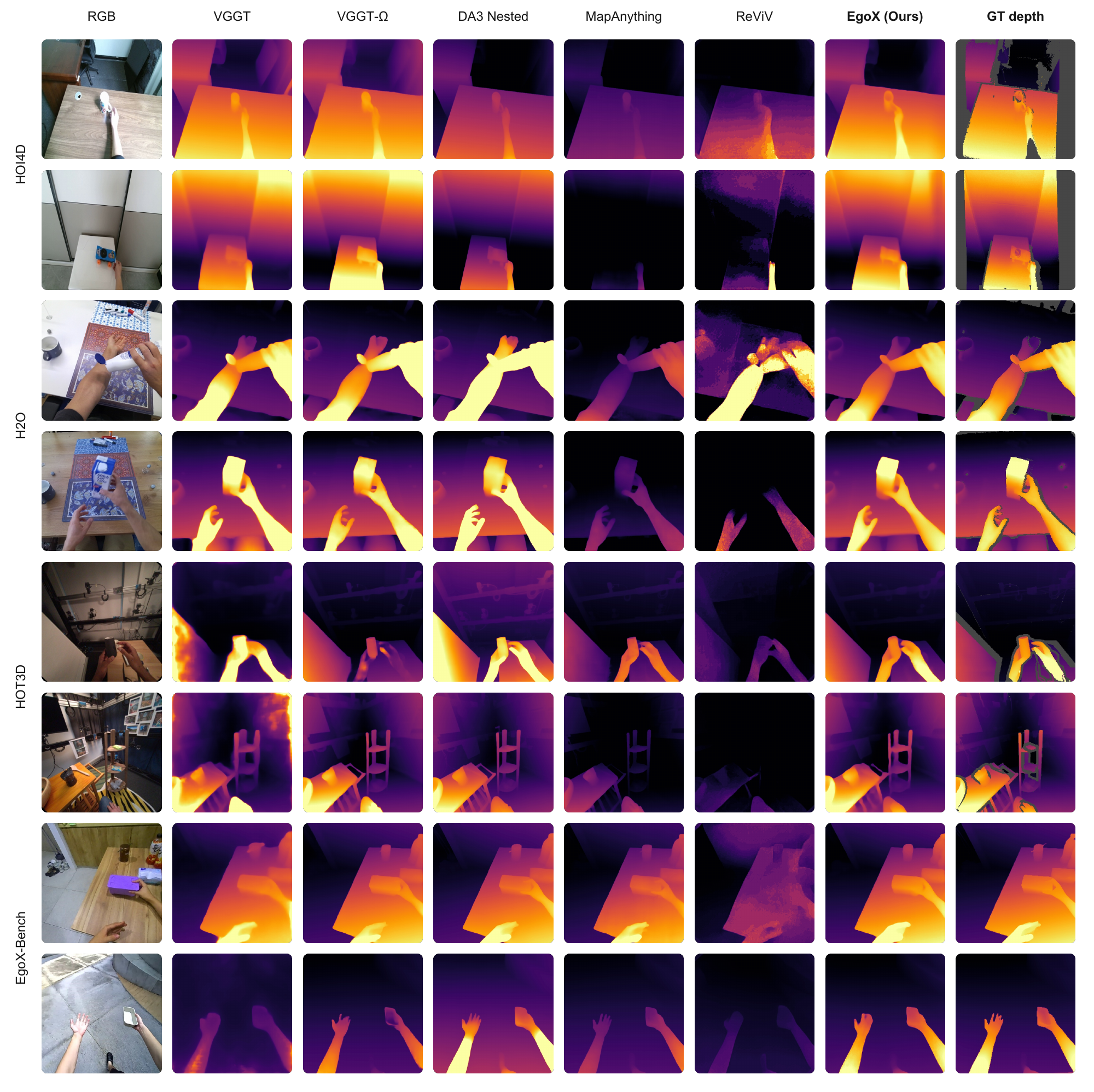}
\caption{\textbf{Additional Depth Reconstruction Results.}
Comparisons use the same reference-depth processing and
scale protocol as Figure~\ref{fig:depth_comparison}.}
\label{fig:depth_comparison_additional}
\end{figure}

Figure~\ref{fig:depth_comparison_additional} supplements
Figure~\ref{fig:depth_comparison} with additional depth predictions.

\section{Training and Compute}
\label{app:training}
\label{sec:method:training}



\paragraph{Training supervision.}
We combine the EgoX-Data-6M training split with six independently
sourced public interaction datasets: HOT3D~\citep{hot3d2024}, HOI4D~\citep{liu2022hoi4d}, H2O~\citep{kwon2021h2o}, HO-Cap~\citep{hocap2024},
H2O-3D~\citep{keypointtransformer2022}, and DexYCB~\citep{chao2021dexycb}. We reprocess them so that depth, hand geometry,
and object geometry agree under consistent metric camera conventions,
avoiding contradictory supervision across tasks. Additional real-world
sources are Xperience-10M~\citep{xperience_10m}, ACE-Data-0~\citep{cao2026ace}, and AgiBot World 2026~\citep{agibotworld2026}; synthetic sources are Aria
Synthetic Environments~\citep{avetisyan2024scenescriptreconstructingscenesautoregressive}, BEHAVIOR-1K~\citep{li2024behavior1k}, SceneNet RGB-D~\citep{McCormac_2017_ICCV}, and Dynamic
Replica~\citep{karaev2023dynamicstereo}. The pool contains approximately 280 hours. Training first
mixes real and synthetic sources and then continues on an equal mixture
of the retained real interaction sources. We sample
$T\in\{1,\ldots,24\}$ observations per input set.

\paragraph{Architecture details.}
Patch features from DINOv3 ViT-L/16 feed a 24-block reconstruction
Transformer with a hidden dimension of 1024. Nineteen blocks exchange
all tokens across views, while the remaining five operate on compact
register and entity tokens. Camera and depth heads read fused image
features, whereas hand and object heads read their view-specific entity
states. The model uses learned spatial and reference embeddings without
temporal encoding.

\paragraph{Coordinate and scale conventions.}
Training uses a common normalization factor $z_{\mathrm{gt}}$, defined
as the mean Euclidean distance of valid metric depth points from the
reference-camera origin after transformation into $\mathcal W$. Metric
depth and translations are divided by $z_{\mathrm{gt}}$; rotations and
intrinsics are unchanged. For wrist-centred MANO joints
$\hat J_{\mathrm{rel},t}^{m}$ and reconstructed wrist $\hat w_t^n$,
training-space and metric joints are
\begin{equation}
    \hat J_t^n
    =
    \frac{\hat J_{\mathrm{rel},t}^{m}}{z_{\mathrm{gt}}}
    +\hat w_t^n,
    \qquad
    \hat J_t^m
    =
    \hat J_{\mathrm{rel},t}^{m}
    +\hat s\,\hat w_t^n.
    \label{eq:mano_metric_composition_app}
\end{equation}
Object translations use the annotated object origin. Object orientation
is represented relative to the reference view as
$\Delta A_{t,j}^o=A_{t,j}^o(A_{r,j}^o)^\top$; the corresponding
camera-conditioned target is
$Q_{t,j}^{o,*}=R_t^*\Delta A_{t,j}^{o,*}$.

\paragraph{Detailed objective.}
The parameter term applies $\ell_1$ losses to normalized camera
translation, standardized unit quaternions, field of view, and
normalized principal point; geodesic losses to hand and object
rotations; Smooth L1 losses to MANO shape, image positions, and
log-depths; and class-weighted binary cross-entropy to in-frame
predictions. Image-position residuals are normalized by the longer
image dimension. Ground-truth cameras define the view-local targets.

The dense losses $\mathcal L_D$ and $\mathcal L_P$ follow
VGGT-$\Omega$, combining confidence-weighted residuals, inverse-depth
weighting, confidence regularization, and multiscale gradient
penalties. Hand and object reconstruction use Smooth L1 penalties
after composition through predicted cameras. Object rotation uses
$d_{\mathrm{sym}}(A,B)=\min_{G\in\mathcal S_r}d(A,BG)$, where $d$ is
geodesic distance and $\mathcal S_r$ is the object's symmetry group in
the reference orientation. The reference rotation is supervised
toward identity without symmetry relaxation; the coupled term applies
to non-reference views. All reductions use annotation-validity and
in-frame masks, and object terms additionally require a bound
reference mask. Missing annotations are not treated as negative labels.

\paragraph{Sampling and optimization.}
The source composition and two-stage mixture are summarized above. Each per-GPU batch samples
$T\in\{1,\ldots,24\}$ and uses a $T$-dependent batch size to maintain
an approximately constant image budget. The image encoder is initialized
from pretrained DINOv3 weights and the remaining modules randomly.
All modules are optimized jointly using AdamW with linear warm-up,
cosine decay, mixed precision, and module-wise gradient clipping.



\end{document}